\documentclass{article}

\usepackage[preprint]{colm2026_conference}

\usepackage[T1]{fontenc}
\usepackage{microtype}
\usepackage{graphicx}
\usepackage{subcaption}
\usepackage{booktabs} %
\usepackage{lineno}

\usepackage{hyperref}

\definecolor{darkblue}{rgb}{0, 0, 0.5}
\hypersetup{colorlinks=true, citecolor=darkblue, linkcolor=darkblue, urlcolor=darkblue}

\usepackage{amsmath}
\usepackage{amssymb}
\usepackage{mathtools}

\usepackage{amsmath,amsfonts,bm}

\def\eqref#1{equation~\ref{#1}}

\def\1{\bm{1}}

\DeclareMathAlphabet{\mathsfit}{\encodingdefault}{\sfdefault}{m}{sl}
\SetMathAlphabet{\mathsfit}{bold}{\encodingdefault}{\sfdefault}{bx}{n}

\usepackage{url}
\usepackage{enumitem}
\usepackage{rotating} %
\usepackage{colortbl} %
\newcommand{\ccol}[1]{\cellcolor[HTML]{#1}} %
\usepackage{multirow} %
\usepackage{makecell} %
\usepackage{tabularx} %
\usepackage{array} %
\usepackage{longtable} %
\usepackage{listings} %
\usepackage{lstautogobble} %
\usepackage{placeins}  %
\usepackage{wrapfig}   %
\usepackage{float}     %
\usepackage{pifont}    %
\newcommand{\cmark}{\ding{51}}  %
\newcommand{\xmark}{\ding{55}}  %
\usepackage[most]{tcolorbox}  %

\graphicspath{{figures/}{icons/}}
\newcommand{\modelicon}[1]{\raisebox{-0.15em}{\includegraphics[height=0.9em]{#1.png}}}
\newcommand{\qwenicon}{\modelicon{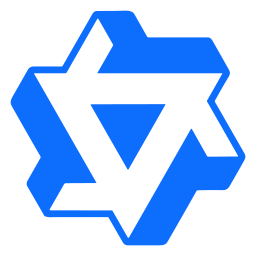}}           %
\newcommand{\openaiicon}{\modelicon{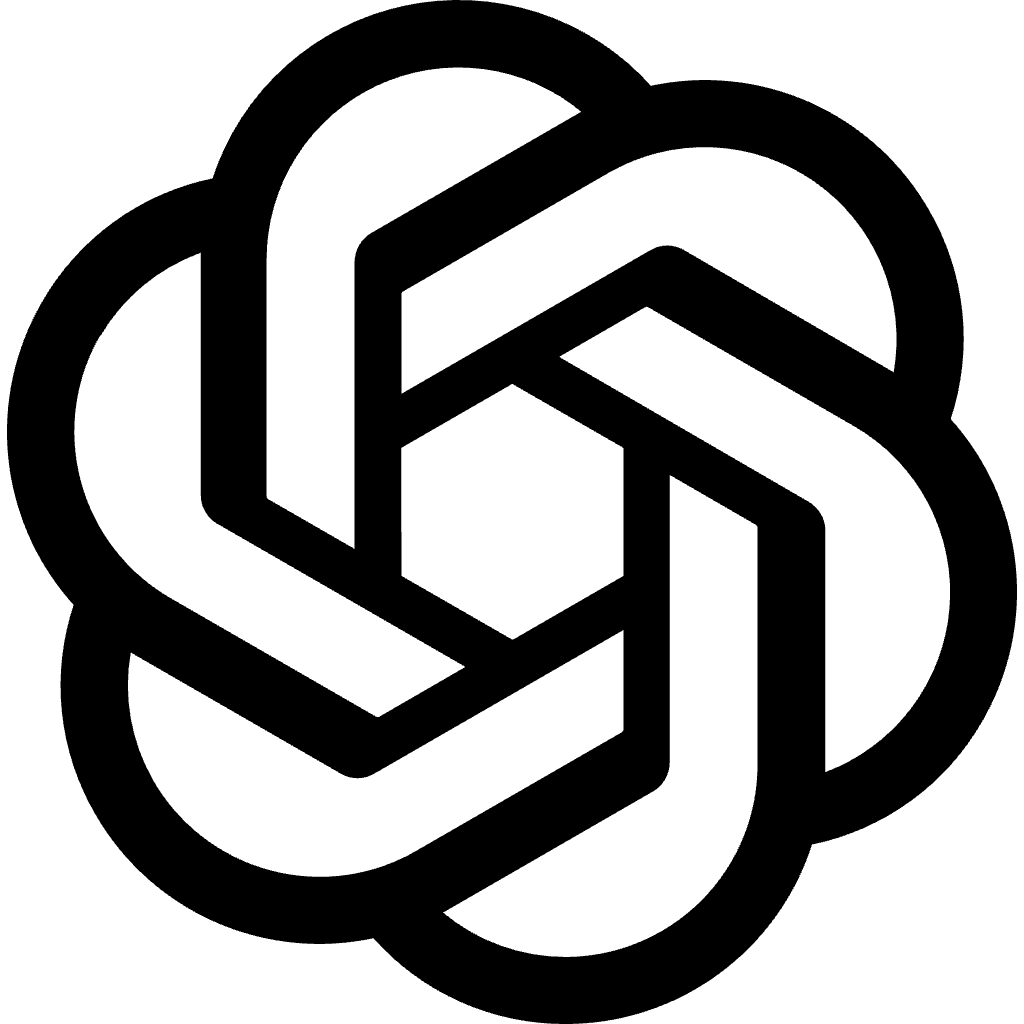}}       %
\newcommand{\anthropicicon}{\modelicon{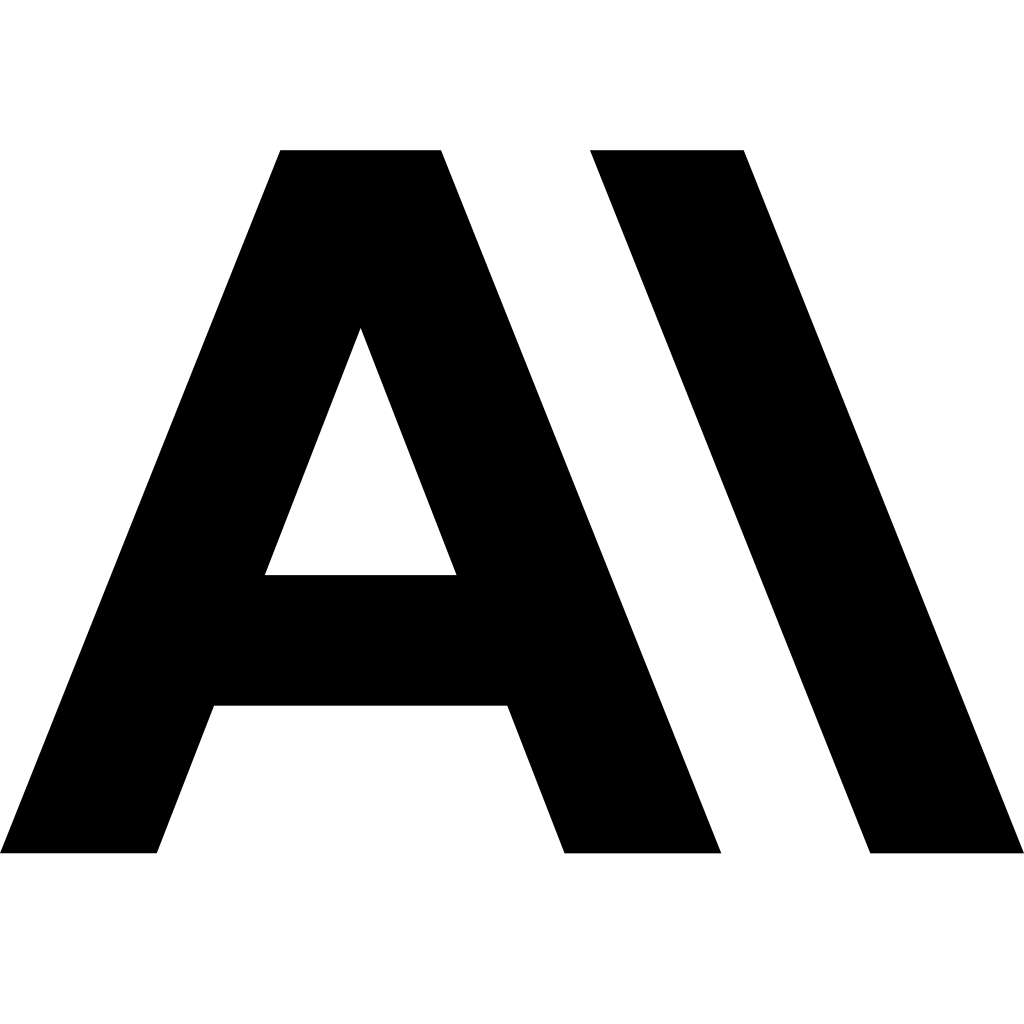}} %
\newcommand{\hficon}{\modelicon{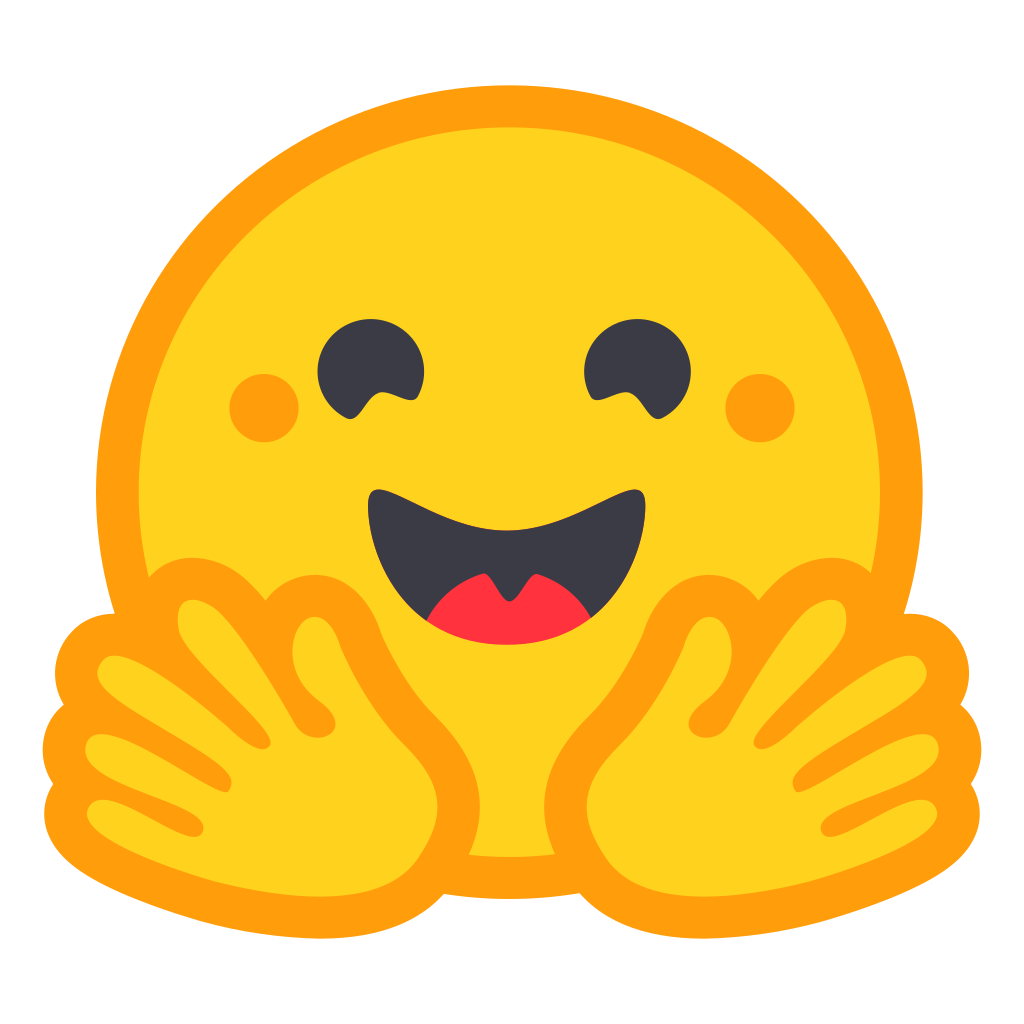}}      %
\newcommand{\allenaiicon}{\modelicon{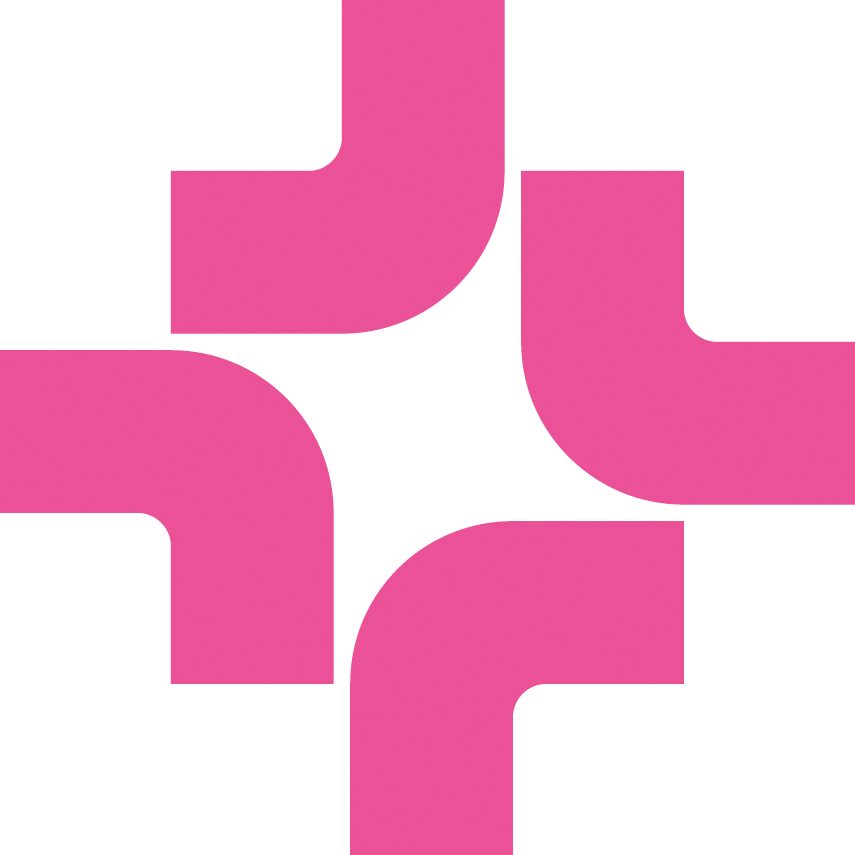}}         %
\newcommand{\nvidiaicon}{\modelicon{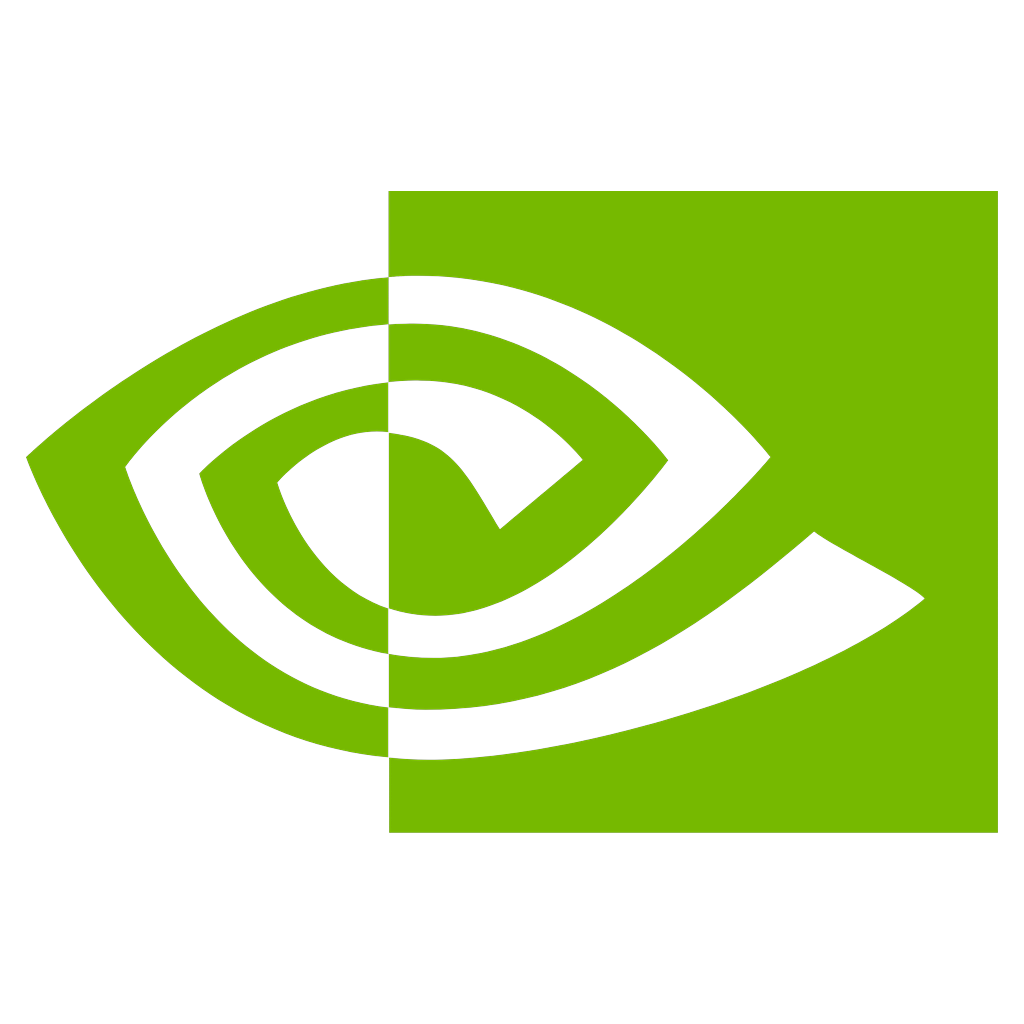}}       %
\newcommand{\xaiicon}{\modelicon{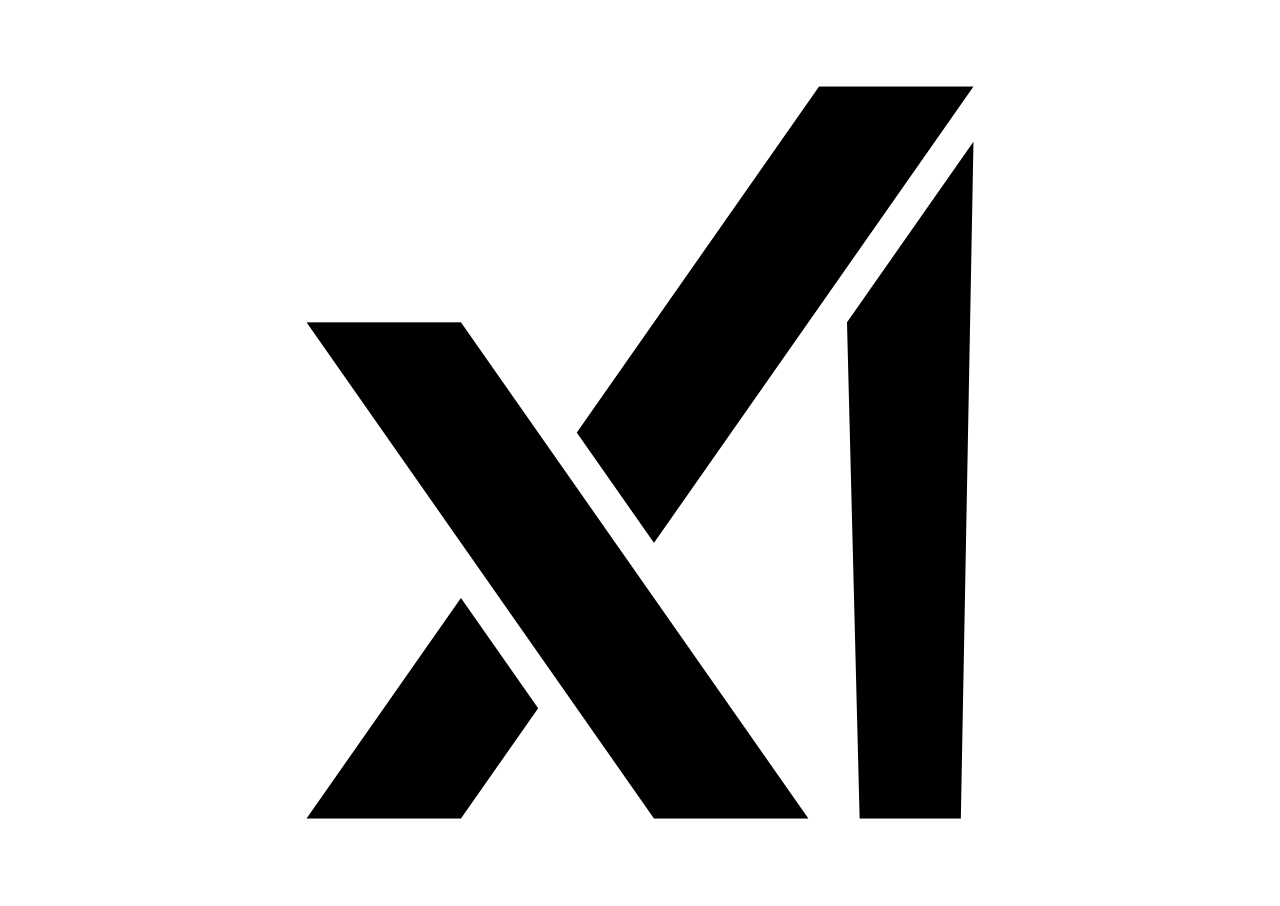}}              %

\newlist{namedlist}{description}{1}
\setlist[namedlist]{
    font=\normalfont\bfseries,
    leftmargin=2em,
    labelindent=0em,
    itemindent=0em
}

\newcommand{\model}[1]{\texttt{#1}}

\title{The Format Tax}

\author{
  Ivan Yee Lee\thanks{Corresponding author: \texttt{iylee@ucsd.edu}} \quad
  Loris D'Antoni \quad
  Taylor Berg-Kirkpatrick \\
  UC San Diego
}

\begin{document}

\maketitle

\begin{abstract}
Asking a large language model to respond in JSON should be a formatting choice, not a capability tax. Yet we find that structured output requirements---JSON, XML, LaTeX, Markdown---substantially degrade reasoning and writing performance across open-weight models. The research response has focused on constrained decoding, but sampling bias accounts for only a fraction of the degradation. The dominant cost enters at the prompt: format-requesting instructions alone cause most of the accuracy loss, before any decoder constraint is applied. This diagnosis points to a simple principle: decouple reasoning from formatting. Whether by generating freeform first and reformatting in a second pass, or by enabling extended thinking within a single generation, separating the two concerns substantially recovers lost accuracy. Across six open-weight models, four API models, four formats, and tasks spanning math, science, logic, and writing, decoupling recovers most lost accuracy. Notably, most recent closed-weight models show little to no format tax, suggesting the problem is not inherent to structured generation but a gap that current open-weight models have yet to close. Code is available at \url{https://github.com/ivnle/the-format-tax}.
\end{abstract}

\section{Introduction}

Structured output has become the default interface between language models and software systems. Function calling, tool use, and agentic frameworks all route model generations through JSON schemas or equivalent format constraints. Grammar-constrained decoding guarantees this by masking invalid tokens during generation, ensuring every output parses. The implicit contract: syntax enforcement should be invisible to reasoning. The model thinks the same thoughts; they simply emerge in a valid format. We show this contract is broken. Requiring structured output degrades reasoning and writing performance across open-weight models---though recent closed-weight models appear to have closed this gap.

The degradation we observe is not new. \citet{tam2024letspeakfreelystudy} documented accuracy drops when models generate structured output, and practitioners have long suspected that format requirements carry hidden costs. What remains unclear is \emph{why}. The research response has targeted the decoder. Grammar-constrained decoding restricts sampling to tokens that satisfy the grammar, but this restriction distorts the model's distribution---biasing it away from what the model would produce unconstrained. Grammar-aligned decoding methods \citep{gad} attempt to correct this. This line of work is principled---but it rests on an assumption: that the decoder is the primary source of degradation. What if sampling bias accounts for only a fraction of the loss?

To answer this, we distinguish two sources of degradation by \emph{when} they occur. The format-requesting prompt introduces one cost before generation begins: the model must now produce reasoning and valid structure in the same output. Constrained decoding introduces a second cost during generation by masking tokens. If the prompt-level cost dominates, no amount of decoder engineering can close the gap. We test this across MATH-500 (competition math), GPQA (graduate science), ZebraLogic (constraint satisfaction), and WritingBench (open-ended writing)---four tasks, four output formats, and ten models spanning open-weight and commercial APIs.

The results confirm that the prompt-level cost dominates. Across most open-weight model--task--format combinations, structured output reduces accuracy compared to freeform generation, whether or not constrained decoding is applied. Standard prompt-level mitigations---few-shot examples and explicit schema descriptions---leave the majority of the degradation intact. What does work is decoupling reasoning from formatting. Letting the model reason freely before producing structured output---either via a second reformatting pass or extended thinking---recovers most of the lost accuracy. Recent closed-weight models show near-zero degradation, suggesting that the tax is not inherent to structured generation. The format tax, it turns out, is paid at the prompt---not the decoder.

In summary, we (1) systematically measure the format tax across ten models, four formats, and tasks spanning reasoning and writing; (2) decompose it into prompt-level degradation (dominant) and decoder-level sampling bias (minor); (3) show that prompt-level mitigations---schema descriptions, few-shot examples---do not close the gap; (4) demonstrate that decoupling reasoning from formatting, via either two-pass generation or extended thinking, recovers most lost accuracy; and (5) show that recent closed-weight models resist the format tax, suggesting the problem is solvable.
\section{Background}
\label{sec:background}

A formal grammar $\mathcal{G} = (V, \Sigma, R, S)$ consists of non-terminal symbols $V$, terminal symbols $\Sigma$, production rules $R$, and a start symbol $S$. It defines a language $\mathcal{L}(\mathcal{G})$: the set of all strings derivable from $S$ by applying rules in $R$. Any string in $\mathcal{L}(\mathcal{G})$ is well-formed by construction.

Grammar-constrained decoding (GCD) masks tokens that would violate a target grammar, guaranteeing syntactically valid output \citep{gcd, willard2023efficientguidedgenerationlarge, park2025flexibleefficientgrammarconstraineddecoding}. At each step, the model's distribution is restricted to valid continuations:
\begin{equation}
\label{eq:gcd_prob}
p_{\text{GCD}}(x_t \mid x_{<t}, \mathcal{G}) \propto p(x_t \mid x_{<t}) \cdot \mathbf{1}[x_t \text{ valid under } \mathcal{G}]
\end{equation}
This guarantee has driven rapid adoption: APIs must return parseable JSON, function-calling agents need well-formed arguments, and downstream systems cannot recover from malformed output. Major frameworks now offer built-in support, including OpenAI's structured outputs, vLLM, Outlines, and Guidance.

Grammar-constrained decoding operates myopically: at each step, it masks tokens that would immediately violate the grammar, without considering whether the remaining valid tokens lead to feasible completions. Grammar-aligned decoding (GAD; \citealt{gad}) methods address this by targeting the true intersection of the language model and the grammar:
\begin{equation}
\label{eq:gad_prob}
p_{\text{GAD}}(x) \propto p(x) \cdot \mathbf{1}[x \in \mathcal{L}(\mathcal{G})]
\end{equation}
Sampling from this distribution exactly is intractable; practical methods approximate it via adaptive sampling \citep{gad}, MCMC \citep{cars, gad_mcmc}, or Sequential Monte Carlo \citep{awrs, genlm_smc}. These methods share a common premise: that the myopia of token-level masking is the primary source of degradation, and that better approximations should recover lost reasoning quality. Rather than benchmark individual GAD implementations, we test this shared premise directly by measuring how much degradation is attributable to the decoder in the first place.

\begin{tcolorbox}[
  colback=gray!8,
  colframe=gray!40,
  boxrule=0.4pt,
  arc=1.5pt,
  left=5pt, right=5pt, top=4pt, bottom=4pt,
  fontupper=\small
]
\textbf{Definition: Format Tax.} Format constraints specify presentation (JSON, XML, etc.), not content---a pure transduction that should leave task performance unchanged. The format tax measures the gap between this expectation and reality:
\vspace{-0.5em}
\begin{equation*}
\text{Format Tax} = \text{Perf}_{\text{freeform}} - \text{Perf}_{\text{format-constrained}}
\end{equation*}
\vspace{-1em}

\noindent A positive value indicates that enforcing format degrades performance. We operationalize performance as task accuracy for reasoning benchmarks and LLM-judged quality for writing tasks.
\end{tcolorbox}

\section{Experimental Setup}
\label{sec:experimental_design}
\begin{table}[t]
\centering
\footnotesize
\begin{tabular*}{\columnwidth}{@{}l@{\extracolsep{\fill}}l@{\hspace{2em}}l@{\extracolsep{\fill}}l@{}}
\rowcolor{gray!15} \multicolumn{4}{@{}l}{\textbf{Freeform}} \\
\texttt{··T·} & $x \sim p(x \mid \texttt{··T})$ & & \\
\rowcolor{gray!15} \multicolumn{2}{@{}l}{\textbf{1-Turn}} & \multicolumn{2}{l}{\textbf{1-Turn + Think}} \\
\texttt{GET·} & $x \sim p(x \mid \texttt{GET})$ & \texttt{GET·} & $r \sim p(r \mid \texttt{GET})$; \; $x \sim p(x \mid r, \texttt{GET})$ \\
\texttt{G·T·} & $x \sim p(x \mid \texttt{G·T})$ & \texttt{G·T·} & $r \sim p(r \mid \texttt{G·T})$; \; $x \sim p(x \mid r, \texttt{G·T})$ \\
\texttt{·ET·} & $x \sim p(x \mid \texttt{·ET})$ & \texttt{·ET·} & $r \sim p(r \mid \texttt{·ET})$; \; $x \sim p(x \mid r, \texttt{·ET})$ \\
\texttt{GETC} & $x \sim p_{\text{GCD}}(x \mid \texttt{GET})$ & \texttt{GETC} & $r \sim p(r \mid \texttt{GET})$; \; $x \sim p_{\text{GCD}}(x \mid r, \texttt{GET})$ \\
\texttt{G·TC} & $x \sim p_{\text{GCD}}(x \mid \texttt{G·T})$ & \texttt{G·TC} & $r \sim p(r \mid \texttt{G·T})$; \; $x \sim p_{\text{GCD}}(x \mid r, \texttt{G·T})$ \\
\texttt{·ETC} & $x \sim p_{\text{GCD}}(x \mid \texttt{·ET})$ & \texttt{·ETC} & $r \sim p(r \mid \texttt{·ET})$; \; $x \sim p_{\text{GCD}}(x \mid r, \texttt{·ET})$ \\
\rowcolor{gray!15} \multicolumn{2}{@{}l}{\textbf{2-Turn}} & \multicolumn{2}{l}{\textbf{2-Turn + Think}} \\
\texttt{GET·} & $a \sim p(a \mid \texttt{··T})$; \; $x \sim p(x \mid a, \texttt{GET})$ & \texttt{GET·} & $r, a \sim p(r, a \mid \texttt{··T})$; \; $x \sim p(x \mid a, \texttt{GET})$ \\
\texttt{G·T·} & $a \sim p(a \mid \texttt{··T})$; \; $x \sim p(x \mid a, \texttt{G·T})$ & \texttt{G·T·} & $r, a \sim p(r, a \mid \texttt{··T})$; \; $x \sim p(x \mid a, \texttt{G·T})$ \\
\texttt{·ET·} & $a \sim p(a \mid \texttt{··T})$; \; $x \sim p(x \mid a, \texttt{·ET})$ & \texttt{·ET·} & $r, a \sim p(r, a \mid \texttt{··T})$; \; $x \sim p(x \mid a, \texttt{·ET})$ \\
\texttt{GETC} & $a \sim p(a \mid \texttt{··T})$; \; $x \sim p_{\text{GCD}}(x \mid a, \texttt{GET})$ & \texttt{GETC} & $r, a \sim p(r, a \mid \texttt{··T})$; \; $x \sim p_{\text{GCD}}(x \mid a, \texttt{GET})$ \\
\texttt{G·TC} & $a \sim p(a \mid \texttt{··T})$; \; $x \sim p_{\text{GCD}}(x \mid a, \texttt{G·T})$ & \texttt{G·TC} & $r, a \sim p(r, a \mid \texttt{··T})$; \; $x \sim p_{\text{GCD}}(x \mid a, \texttt{G·T})$ \\
\texttt{·ETC} & $a \sim p(a \mid \texttt{··T})$; \; $x \sim p_{\text{GCD}}(x \mid a, \texttt{·ET})$ & \texttt{·ETC} & $r, a \sim p(r, a \mid \texttt{··T})$; \; $x \sim p_{\text{GCD}}(x \mid a, \texttt{·ET})$ \\
\end{tabular*}
\caption{Experimental configurations organized by generation strategy and prompt variation. GETC notation: \textbf{G}rammar, \textbf{E}xamples, \textbf{T}ask, \textbf{C}onstrained decoding. G, E, and T vary prompt contents; C activates grammar enforcement during decoding. Dots indicate absence. $p$ samples without constraints; $p_{\text{GCD}}$ masks tokens that violate the grammar at each decoding step. Thinking variants first sample reasoning $r$, then formatted output $x$. For 2-Turn, Turn~1 produces answer $a$ (task only) while Turn~2 formats.}
\label{tab:strategies}
\end{table}

Our experimental design isolates the source of degradation by varying two dimensions while holding task and format constant: \emph{generation strategy} (how reasoning and formatting are organized) and \emph{prompt contents} (what format guidance the model receives). Table~\ref{tab:strategies} summarizes five strategies ranging from standard single-pass generation to 2-Turn pipelines that separate reasoning from formatting into distinct calls---Turn~1 answers in natural language, Turn~2 reformats under constraints. Thinking-enabled models generate a hidden reasoning trace (a ``scratchpad'') before producing their visible output, separating internal deliberation from the final answer.

These dimensions yield two critical tests. First, comparing 1-Turn with and without grammar constraints, holding the prompt constant, isolates the decoder: if degradation persists without GCD, the tax is paid upstream. Second, if separating reasoning from formatting---via either 2-Turn or thinking---recovers accuracy despite the same format constraints, the tax is not an unavoidable cost of structured output. We also systematically vary prompt contents to test whether richer format guidance closes the gap: if the format tax were a prompting deficiency, enriching the prompt should eliminate it.

\subsection{Prompt Variations}

We vary prompt contents using a four-letter notation: \textbf{G} (grammar shown), \textbf{E} (few-shot examples), \textbf{T} (task description, always present), and \textbf{C} (constrained decoding active). G, E, and T vary what appears in the prompt; C activates grammar enforcement during decoding. Dots indicate absence; \texttt{GET·} provides grammar, examples, and task but no constraint, while \texttt{G·TC} includes grammar and constraint but no examples.

Without format information in the prompt, \texttt{··T·} reduces to freeform generation---no structure is requested or enforced---and serves as our baseline. We exclude \texttt{··TC}: applying grammar constraints without any format guidance in the prompt produces incoherent output.

\subsection{Models, Tasks, and Formats}

We evaluate ten models. Six open-weight models (\model{qwen3-32b}, \model{qwen3-8b}, \model{olmo3.1-32b}, \model{olmo3-7b}, \model{smollm3-3b}, \model{nemotron3-nano}) run locally via vLLM; four API models (\model{gpt-5-nano}, \model{gpt-5.4-nano}, \model{claude-haiku-4.5}, \model{grok-4.1-fast}) use native structured output endpoints. All are instruction-tuned, 2024--2025 releases.

Our tasks span reasoning and writing. \textbf{MATH-500} \citep{hendrycks2021math}: 500 competition math problems requiring multi-step symbolic reasoning with open-ended answers. \textbf{GPQA-Diamond} \citep{rein2023gpqa}: 198 graduate-level science questions (physics, chemistry, biology) in 4-choice format. \textbf{ZebraLogic} \citep{lin2025zebralogic}: 500 constraint satisfaction puzzles requiring systematic deduction over multiple entities and attributes, in multiple-choice format (2--6 options). \textbf{WritingBench} \citep{wu2025writingbench}: 500 open-ended writing tasks judged by LLM on prose quality.

For formats, we test \textbf{JSON Schema}, \textbf{XML}, \textbf{Markdown}, and \textbf{LaTeX}---spanning both structured data formats common in API settings and document markup increasingly required in agent-generated content. All four are compared against a \textbf{freeform} baseline with no format requirements. Appendix~\ref{sec:format-constraints} shows representative constraints and example outputs.

\subsection{Evaluation}

Our goal is to measure reasoning quality independent of format compliance. For each question, we first attempt to parse the structured output; if parsing fails, we fall back to task-specific regex extraction. For GPQA and ZebraLogic, this pipeline is reliable because answers are multiple-choice letters. For MATH-500, we added symbolic normalization via SymPy but found that correct answers embedded in different format markup were still systematically missed, inflating the measured tax. We therefore score MATH-500 with a fixed LLM judge (\model{gpt-5.4-nano}) that compares the model's response against gold for mathematical equivalence, which \emph{reduces} the measured tax relative to deterministic extraction (Appendix~\ref{sec:math500_scoring}).

WritingBench also uses an LLM judge (\model{gpt-5.2}), but because there is no gold-standard answer. A concern is that the judge might penalize LaTeX markup rather than content. To control for this, we strip markup from each response and judge both the original and the extracted prose. If scores are similar, the degradation reflects genuinely worse writing; if prose scores higher, the judge was biased by formatting.

\section{Results}
\label{sec:results}

Our experiments establish that the format tax is widespread among open-weight models but largely absent in recent closed-weight models, that its dominant source is the format-requesting prompt rather than grammar-constrained decoding, and that separating the task from its formatting recovers most lost accuracy. Full per-configuration results appear in Appendix~Table~\ref{tab:format-tax-consolidated}.

\subsection{Does the Format Tax Exist?}
\label{sec:format_tax}

\paragraph{Writing quality.}
Requesting LaTeX output directly hurts writing quality substantially (Table~\ref{tab:writingbench}). A natural concern is that LLM judges penalize markup aesthetics rather than content. We tested this by stripping LaTeX and judging the extracted prose: scores did not meaningfully change. The degradation reflects genuinely worse writing, not judge bias.
Wilcoxon signed-rank tests confirm the effect is statistically significant in 36 of 40 cells ($p < 0.05$).


\begin{table}[htbp]
\centering
\small
\setlength{\tabcolsep}{3pt}
\resizebox{\textwidth}{!}{%
\begin{tabular}{@{}l rrrrr@{\hspace{1em}}rrrrr@{}}
\toprule
 & \multicolumn{5}{c}{No Thinking} & \multicolumn{5}{c}{With Thinking} \\
\cmidrule(lr){2-6} \cmidrule(lr){7-11}
 & & \multicolumn{2}{c}{1-Turn} & \multicolumn{2}{c}{2-Turn} & & \multicolumn{2}{c}{1-Turn} & \multicolumn{2}{c}{2-Turn} \\
\cmidrule(lr){3-4} \cmidrule(lr){5-6} \cmidrule(lr){8-9} \cmidrule(lr){10-11}
Model & Free & LTX & Prose & LTX & Prose & Free & LTX & Prose & LTX & Prose \\
\midrule
\nvidiaicon~nemotron3-nano & 62.3 & \ccol{E99797}-15.7 & \ccol{E89494}-16.1 & \ccol{F5D0D0}-7.0 & \ccol{F5D1D1}-6.9 & 66.4 & \ccol{E58484}-18.6 & \ccol{E68C8C}-17.4 & \ccol{F5CFCF}-7.1 & \ccol{F6D4D4}-6.4 \\
\qwenicon~qwen3-32b & 56.7 & \ccol{F5D1D1}-6.8 & \ccol{F4CBCB}-7.8 & -0.5 & -0.9 & 56.6 & \ccol{F3C7C7}-8.4 & \ccol{F1C0C0}-9.4 & \ccol{FBEDED}-2.6 & \ccol{FBEDED}-2.6 \\
\qwenicon~qwen3-8b & 51.5 & \ccol{EBA4A4}-13.7 & \ccol{F0BCBC}-10.1 & \ccol{F6D6D6}-6.1 & \ccol{FBEEEE}-2.5 & 54.2 & \ccol{EEB0B0}-11.9 & \ccol{F2C3C3}-9.0 & \ccol{F9E4E4}-4.1 & \ccol{FAE8E8}-3.4 \\
\hficon~smollm3-3b & 32.4 & \ccol{F5D1D1}-6.8 & \ccol{F7D9D9}-5.6 & \ccol{F9E3E3}-4.2 & \ccol{F9E3E3}-4.2 & 35.6 & \ccol{F5D3D3}-6.5 & \ccol{F6D4D4}-6.4 & \ccol{ECA9A9}-12.9 & \ccol{ECA9A9}-12.9 \\
\openaiicon~gpt-5-nano & 61.7 & \ccol{F8DFDF}-4.8 & \ccol{F9E4E4}-4.0 & \ccol{FCF5F5}-1.5 & -0.5 & 71.1 & \ccol{F3CACA}-7.9 & \ccol{F6D4D4}-6.4 & \ccol{FDF7F7}-1.1 & -0.1 \\
\bottomrule
\end{tabular}}%
\caption{LaTeX formatting degrades writing quality---but 2-Turn recovers most of the loss.
Freeform shows absolute quality (\%); other columns show change vs.\ freeform.
\textbf{1-Turn} (\texttt{GET·}): model generates LaTeX directly with grammar and examples in the prompt, no constrained decoding.
\textbf{2-Turn} (\texttt{··T·} $\to$ \texttt{GET·}): model generates freeform, then reformats to LaTeX.
Column labels: \textsc{LTX} = judge evaluates document as-is; \textsc{Prose} = markup stripped, evaluating content only.}
\label{tab:writingbench}
\end{table}

\begin{table*}[!ht]
\centering
\small
\setlength{\tabcolsep}{2.5pt}
\resizebox{\textwidth}{!}{%
\begin{tabular}{@{}l *{4}{r}@{\hspace{0.6em}}*{4}{r}@{\hspace{0.6em}}*{4}{r}@{\hspace{0.8em}}r@{}}
\toprule
 & \multicolumn{4}{c}{GPQA} & \multicolumn{4}{c}{Math} & \multicolumn{4}{c}{Zebra} & \\
\cmidrule(lr){2-5} \cmidrule(lr){6-9} \cmidrule(lr){10-13}
Model & JSON & XML & MD & LTX & JSON & XML & MD & LTX & JSON & XML & MD & LTX & \textbf{Avg} \\
\midrule
\rowcolor{gray!15} \qwenicon~qwen3-32b & \multicolumn{4}{c}{52.0} & \multicolumn{4}{c}{84.4} & \multicolumn{4}{c}{64.8} & \textbf{67.1} \\
\quad \texttt{+ GCD} & \ccol{F7DADA}-7.4 & \ccol{F6D7D7}-7.9 & \ccol{F7DCDC}-6.9 & \ccol{F8E1E1}-5.9 & \ccol{F1BFBF}-12.7 & \ccol{F3C8C8}-10.9 & \ccol{F4CCCC}-10.0 & \ccol{FCF1F1}-2.8 & \ccol{F7DBDB}-7.1 & \ccol{F7DADA}-7.3 & \ccol{F4CCCC}-10.1 & \ccol{F7DCDC}-6.9 & \textbf{\ccol{F6D7D7}-8.0} \\
\quad \texttt{no GCD} & \ccol{F7DBDB}-7.1 & \ccol{F7DADA}-7.4 & \ccol{FBECEC}-3.7 & \ccol{F9E5E5}-5.1 & \ccol{F7DCDC}-6.8 & \ccol{F8DEDE}-6.5 & \ccol{FAEAEA}-4.2 & \ccol{FAEAEA}-4.0 & \ccol{F8DEDE}-6.6 & \ccol{F7DADA}-7.3 & \ccol{F8E2E2}-5.8 & \ccol{F9E4E4}-5.3 & \textbf{\ccol{F8E2E2}-5.8} \\

\rowcolor{gray!15} \qwenicon~qwen3-8b & \multicolumn{4}{c}{40.6} & \multicolumn{4}{c}{84.0} & \multicolumn{4}{c}{63.1} & \textbf{62.5} \\
\quad \texttt{+ GCD} & \ccol{F4CBCB}-10.3 & \ccol{F8DFDF}-6.4 & \ccol{FBEFEF}-3.0 & \ccol{FDF7F7}-1.5 & \ccol{EEB1B1}-15.6 & \ccol{F5D0D0}-9.3 & \ccol{F6D6D6}-8.2 & \ccol{F9E4E4}-5.3 & \ccol{F4CBCB}-10.4 & \ccol{F6D7D7}-7.9 & \ccol{F5D3D3}-8.8 & \ccol{ECA7A7}-17.5 & \textbf{\ccol{F5D3D3}-8.7} \\
\quad \texttt{no GCD} & \ccol{F7D9D9}-7.6 & \ccol{FCF1F1}-2.7 & \ccol{FAEAEA}-4.0 & \ccol{FDF7F7}-1.5 & \ccol{F5D1D1}-9.1 & \ccol{FAE7E7}-4.6 & \ccol{FAE9E9}-4.3 & \ccol{F9E6E6}-4.8 & \ccol{F6D5D5}-8.3 & \ccol{F5D4D4}-8.6 & \ccol{F7DADA}-7.4 & \ccol{EDABAB}-16.8 & \textbf{\ccol{F7DDDD}-6.6} \\

\rowcolor{gray!15} \allenaiicon~olmo3.1-32b & \multicolumn{4}{c}{53.7} & \multicolumn{4}{c}{91.4} & \multicolumn{4}{c}{72.9} & \textbf{72.7} \\
\quad \texttt{+ GCD} & \ccol{F8E2E2}-5.7 & \ccol{FAEAEA}-4.0 & +0.7 & -0.7 & \ccol{FAEBEB}-3.9 & \ccol{F2C3C3}-12.0 & \ccol{F8E0E0}-6.1 & \ccol{FAE7E7}-4.6 & \ccol{E2F2E2}+2.0 & \ccol{FDF9F9}-1.2 & \ccol{EAF5EA}+1.5 & \ccol{9DD39D}+6.9 & \textbf{\ccol{FCF3F3}-2.3} \\
\quad \texttt{no GCD} & \ccol{FAE7E7}-4.7 & \ccol{FDF7F7}-1.5 & -0.2 & \ccol{FCF2F2}-2.5 & \ccol{FAE7E7}-4.7 & \ccol{FAEAEA}-4.1 & \ccol{FAE9E9}-4.3 & \ccol{FAEAEA}-4.2 & \ccol{D0EAD0}+3.3 & \ccol{B2DCB2}+5.5 & +0.9 & \ccol{B3DDB3}+5.4 & \textbf{-0.9} \\

\rowcolor{gray!15} \allenaiicon~olmo3-7b & \multicolumn{4}{c}{43.3} & \multicolumn{4}{c}{90.5} & \multicolumn{4}{c}{69.3} & \textbf{67.7} \\
\quad \texttt{+ GCD} & \ccol{F3C9C9}-10.8 & \ccol{F5D3D3}-8.8 & -0.8 & \ccol{FBECEC}-3.7 & \ccol{F4CECE}-9.8 & \ccol{F0BBBB}-13.6 & \ccol{FAE8E8}-4.4 & \ccol{F7DDDD}-6.7 & \ccol{F8DEDE}-6.6 & \ccol{F7DBDB}-7.2 & \ccol{FDF7F7}-1.6 & \ccol{F7DBDB}-7.0 & \textbf{\ccol{F7DDDD}-6.7} \\
\quad \texttt{no GCD} & \ccol{F6D5D5}-8.4 & \ccol{FAEBEB}-3.9 & -0.3 & \ccol{F9E5E5}-5.1 & \ccol{F8DFDF}-6.3 & \ccol{F8E2E2}-5.7 & \ccol{FCF4F4}-2.0 & \ccol{FBECEC}-3.8 & \ccol{FDF8F8}-1.3 & \ccol{F7DBDB}-7.2 & \ccol{E5F3E5}+1.8 & -0.2 & \textbf{\ccol{FBEDED}-3.5} \\

\rowcolor{gray!15} \hficon~smollm3-3b & \multicolumn{4}{c}{22.9} & \multicolumn{4}{c}{67.4} & \multicolumn{4}{c}{38.4} & \textbf{42.9} \\
\quad \texttt{+ GCD} & \ccol{B8DFB8}+5.1 & \ccol{FCF2F2}-2.5 & \ccol{D4EBD4}+3.0 & \ccol{84C884}+8.8 & \ccol{E78D8D}-22.8 & \ccol{F0BCBC}-13.4 & \ccol{FBECEC}-3.7 & \ccol{F3C7C7}-11.2 & \ccol{FAE8E8}-4.6 & \ccol{F7DCDC}-6.9 & \ccol{F8E2E2}-5.7 & \ccol{FBEFEF}-3.1 & \textbf{\ccol{F9E7E7}-4.8} \\
\quad \texttt{no GCD} & +0.3 & \ccol{E7F4E7}+1.7 & \ccol{D6EDD6}+2.9 & \ccol{B3DDB3}+5.4 & \ccol{F2C3C3}-11.9 & \ccol{F8E1E1}-6.0 & \ccol{FDF8F8}-1.4 & \ccol{F7DDDD}-6.7 & \ccol{F8E0E0}-6.1 & \ccol{F4CECE}-9.7 & \ccol{F8E1E1}-6.0 & \ccol{FAECEC}-3.8 & \textbf{\ccol{FBEDED}-3.4} \\

\rowcolor{gray!15} \nvidiaicon~nemotron3-nano & \multicolumn{4}{c}{53.9} & \multicolumn{4}{c}{71.7} & \multicolumn{4}{c}{43.1} & \textbf{56.2} \\
\quad \texttt{+ GCD} & \ccol{EFB5B5}-14.8 & \ccol{F4CFCF}-9.6 & \ccol{F7DADA}-7.4 & \ccol{F1BDBD}-13.1 & -0.9 & \ccol{FCF1F1}-2.7 & \ccol{FDF7F7}-1.5 & \ccol{FAEAEA}-4.2 & \ccol{A3D5A3}+6.5 & \ccol{C2E3C2}+4.3 & \ccol{F8DFDF}-6.3 & \ccol{FDF5F5}-1.9 & \textbf{\ccol{FAE9E9}-4.3} \\
\quad \texttt{no GCD} & \ccol{F0B8B8}-14.1 & \ccol{E9F5E9}+1.5 & \ccol{F8E1E1}-5.9 & \ccol{FDF5F5}-1.9 & -1.0 & \ccol{F0B8B8}-14.2 & \ccol{FDF6F6}-1.7 & \ccol{E58484}-24.7 & \ccol{88CA88}+8.4 & \ccol{AEDAAE}+5.7 & \ccol{F7DDDD}-6.7 & \ccol{FAE7E7}-4.7 & \textbf{\ccol{F9E6E6}-4.9} \\
\bottomrule
\end{tabular}}%
\caption{Format tax across open-weight models: accuracy change from freeform baseline (\%).
\texttt{+ GCD} averages over constrained configurations (\texttt{GETC}, \texttt{G·TC}, \texttt{·ETC});
\texttt{no GCD} averages over unconstrained configurations (\texttt{GET·}, \texttt{G·T·}, \texttt{·ET·}).
All results: 1-Turn, thinking off.}
\label{tab:format_tax_universal_ow}
\end{table*}

\paragraph{Reasoning benchmarks (open-weight models).}
The degradation extends beyond writing. Table~\ref{tab:format_tax_universal_ow} presents results across six open-weight models, three reasoning benchmarks, and four output formats. Averaged across tasks and formats, every open-weight model shows accuracy decreases relative to freeform generation. Formats do not form a consistent hierarchy---relative severity varies by model and task.

To assess statistical significance, we apply McNemar's test to per-question paired outcomes, comparing each question's freeform result against its structured-output result. Each (model, task, format, GCD) combination defines a cell; across six open-weight models this yields 144 cells. Of these, 76 show statistically significant effects ($p < 0.05$): 72 in the tax direction and only 4 in the credit direction. The effect is most prevalent on MATH-500 (41 of 48 significant, 40 tax) and ZebraLogic (28 of 48, 26 tax), while GPQA shows mostly non-significant effects (7 of 48). The tax varies by model: \model{qwen3-8b} shows significant degradation in 17 of 24 cells ($-$9.9\,pp average), while \model{nemotron3-nano} shows significant effects in 9 of 24 ($-$4.3\,pp average).

A small number of cells show improvements, visible as green entries in the heatmap. These trace to model-specific artifacts---floor effects, degenerate freeform outputs, and format-induced strategy redirection---rather than a systematic benefit of structured output (Appendix~\ref{sec:positive_cells}).

\begin{table*}[!ht]
\centering
\small
\setlength{\tabcolsep}{2.5pt}
\resizebox{\textwidth}{!}{%
\begin{tabular}{@{}l *{4}{r}@{\hspace{0.6em}}*{4}{r}@{\hspace{0.6em}}*{4}{r}@{\hspace{0.8em}}r@{}}
\toprule
 & \multicolumn{4}{c}{GPQA} & \multicolumn{4}{c}{Math} & \multicolumn{4}{c}{Zebra} & \\
\cmidrule(lr){2-5} \cmidrule(lr){6-9} \cmidrule(lr){10-13}
Model & JSON & XML & MD & LTX & JSON & XML & MD & LTX & JSON & XML & MD & LTX & \textbf{Avg} \\
\midrule
\rowcolor{gray!15} \openaiicon~gpt-5-nano & \multicolumn{4}{c}{47.0} & \multicolumn{4}{c}{88.2} & \multicolumn{4}{c}{50.4} & \textbf{61.9} \\
\quad \texttt{+ GCD} & \ccol{FEFBFB}-1.5 & \ccol{F9E4E4}-10.6 & \ccol{F9E3E3}-11.1 & \ccol{F9E3E3}-11.1 & +0.7 & \ccol{FEFBFB}-1.6 & \ccol{E0F1E0}+3.3 & \ccol{FBF0F0}-5.8 & \ccol{FAE8E8}-9.2 & \ccol{F6D7D7}-16.2 & \ccol{F7DDDD}-13.8 & \ccol{F5CFCF}-19.2 & \textbf{\ccol{FAEBEB}-8.0} \\
\quad \texttt{no GCD} & \ccol{FCF2F2}-4.9 & +0.5 & \ccol{FDF8F8}-2.5 & \ccol{FCF4F4}-4.2 & \ccol{FEFBFB}-1.5 & \ccol{FEFAFA}-1.9 & +0.3 & \ccol{FDF5F5}-3.7 & \ccol{F9E3E3}-11.3 & \ccol{F7DCDC}-13.9 & \ccol{F7DADA}-14.7 & \ccol{FCF4F4}-4.2 & \textbf{\ccol{FCF2F2}-5.2} \\

\rowcolor{gray!15} \openaiicon~gpt-5.4-nano & \multicolumn{4}{c}{58.1} & \multicolumn{4}{c}{88.9} & \multicolumn{4}{c}{65.4} & \textbf{70.8} \\
\quad \texttt{+ GCD} & \ccol{FDFAFA}-2.0 & \ccol{F6D7D7}-16.2 & \ccol{F8E1E1}-12.1 & \ccol{F8E1E1}-12.1 & \ccol{EAF6EA}+2.2 & \ccol{E58484}-59.4 & \ccol{FAE9E9}-8.9 & \ccol{F0B8B8}-28.7 & \ccol{D6EDD6}+4.4 & \ccol{F2C2C2}-24.6 & \ccol{F7DADA}-14.8 & \ccol{F5D2D2}-18.2 & \textbf{\ccol{F6D7D7}-15.9} \\
\quad \texttt{no GCD} & \ccol{F4FAF4}+1.2 & \ccol{FCF4F4}-4.4 & \ccol{FEFAFA}-1.7 & -0.5 & -0.5 & \ccol{FDF8F8}-2.7 & -0.6 & \ccol{FDF9F9}-2.3 & \ccol{84C884}+13.5 & \ccol{BBE0BB}+7.4 & \ccol{B6DEB6}+7.9 & \ccol{89CA89}+12.9 & \textbf{\ccol{E8F4E8}+2.5} \\

\rowcolor{gray!15} \anthropicicon~claude-haiku-4.5 & \multicolumn{4}{c}{60.1} & \multicolumn{4}{c}{91.8} & \multicolumn{4}{c}{83.8} & \textbf{78.6} \\
\quad \texttt{+ GCD} & \ccol{F1F8F1}+1.5 & — & — & — & \ccol{E3F2E3}+3.0 & — & — & — & +0.4 & — & — & — & \textbf{\ccol{EFF8EF}+1.6} \\
\quad \texttt{no GCD} & \ccol{ECF6EC}+2.0 & \ccol{EAF6EA}+2.2 & 0.0 & \ccol{F2F9F2}+1.3 & \ccol{F4FAF4}+1.1 & \ccol{F4FAF4}+1.1 & +0.9 & \ccol{F5FAF5}+1.1 & \ccol{F0F8F0}+1.5 & -0.2 & -0.3 & \ccol{E1F1E1}+3.3 & \textbf{\ccol{F4FAF4}+1.2} \\

\rowcolor{gray!15} \xaiicon~grok-4.1-fast & \multicolumn{4}{c}{67.7} & \multicolumn{4}{c}{78.3} & \multicolumn{4}{c}{72.8} & \textbf{72.9} \\
\quad \texttt{+ GCD} & +0.5 & — & — & — & \ccol{D4EBD4}+4.7 & — & — & — & \ccol{B9DFB9}+7.6 & — & — & — & \textbf{\ccol{D8EDD8}+4.3} \\
\quad \texttt{no GCD} & -0.5 & \ccol{F4FAF4}+1.2 & -0.3 & \ccol{FCF5F5}-3.9 & \ccol{EBF6EB}+2.1 & \ccol{CCE8CC}+5.6 & \ccol{DEF0DE}+3.5 & \ccol{F2F9F2}+1.4 & \ccol{DAEEDA}+4.0 & \ccol{D8EDD8}+4.2 & \ccol{92CE92}+11.9 & \ccol{F4FAF4}+1.1 & \textbf{\ccol{E7F4E7}+2.5} \\
\bottomrule
\end{tabular}}%
\caption{Format tax across closed-weight models: accuracy change from freeform baseline (\%).
\texttt{+ GCD} averages over constrained configurations (\texttt{GETC}, \texttt{G·TC}, \texttt{·ETC});
\texttt{no GCD} averages over unconstrained configurations (\texttt{GET·}, \texttt{G·T·}, \texttt{·ET·}).
For non-OpenAI API models, constrained decoding is available only for JSON.
All results: 1-Turn, thinking off.}
\label{tab:format_tax_universal_api}
\end{table*}

\paragraph{Reasoning benchmarks (closed-weight models).}
The picture differs for recent closed-weight models (Table~\ref{tab:format_tax_universal_api}). \model{gpt-5-nano}, an older model, still suffers the format tax ($-$5.2\,pp average without GCD, $-$8.0\,pp with). However, three newer API models---\model{claude-haiku-4.5}, \model{grok-4.1-fast}, and \model{gpt-5.4-nano}---show near-zero or positive deltas without GCD across all tasks and formats, with especially large gains on ZebraLogic. One exception: \model{gpt-5.4-nano}'s \texttt{+GCD} row shows severe degradation on non-JSON formats, a model-specific regression under grammar enforcement that we analyze in Appendix~\ref{sec:api_gcd_collapse}. Paired analysis confirms these gains reflect genuine accuracy improvements, not scoring artifacts (Appendix~\ref{sec:positive_cells}). These results suggest the format tax is not an inherent limitation of structured generation but a gap that can be closed. Since the tax persists in open-weight models, the remainder of our analysis (Sections~\ref{sec:decomposition} and \ref{sec:two_turn}) focuses on understanding its source and mitigation in that setting.

\subsection{What is the Source of the Format Tax?}
\label{sec:decomposition}

The prevailing explanation targets grammar-constrained decoding: masking tokens during generation distorts the sampling process. Our experimental design tests this by comparing \texttt{GET·} (format-requesting prompt, no constraint) against \texttt{GETC} (same prompt, grammar constraint active). Both share the same prompt; only the decoder differs.

\begin{figure}[t]
\centering
\includegraphics[width=\columnwidth]{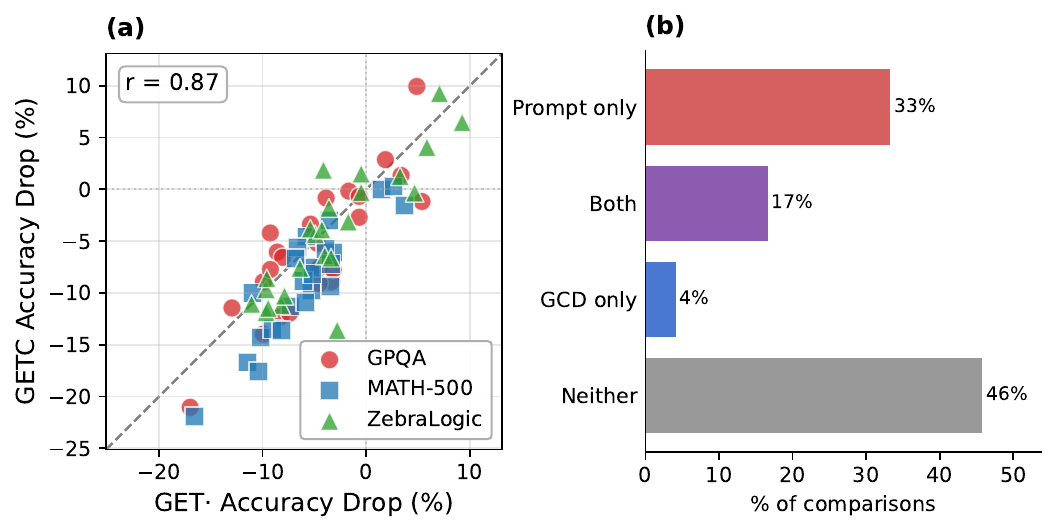}
\caption{Source attribution of the format tax. (a)~Each point is a (model, task, format) combination; the x-axis shows accuracy drop from the format-requesting prompt alone (\texttt{GET·}), while the y-axis adds grammar-constrained decoding (\texttt{GETC}). Points near the diagonal indicate GCD adds little beyond the prompt effect. (b)~Parallel McNemar tests ($p < 0.05$) on 72 cells from open-weight models classify each by whether degradation is already present under the format-requesting prompt alone, added by GCD, or both.}
\label{fig:decomposition}
\end{figure}

If GCD were the culprit, \texttt{GETC} would underperform \texttt{GET·} substantially. Instead, degradation is similar whether or not the constraint is applied (Figure~\ref{fig:decomposition}a). McNemar tests on 72 model--task--format cells (6 models $\times$ 3 tasks $\times$ 4 formats) confirm this asymmetry (Figure~\ref{fig:decomposition}b): of 39 cells with any statistically significant effect, 36 (92\%) are already present under the format-requesting prompt alone and only 15 (38\%) involve GCD. Just 3 cells show GCD-only degradation. The average \texttt{GET·} delta is $-$3.9\,pp, compared to $-$1.6\,pp for GCD.

\paragraph{Corroborating evidence.}
Per-question flip rates---the rate at which freeform-correct answers become incorrect under structured output---are 25.9\% with GCD and 24.1\% without, and the relationship between token compression and flip rate follows the same U-shaped pattern under both conditions (Figure~\ref{fig:flip_rate}; Section~\ref{sec:token_compression} examines this in detail). Separately, grammar-constrained decoding dramatically improves format compliance from 55.7\% to 92.2\% yet accuracy is unchanged---57.3\% without GCD and 55.7\% with, versus 61.5\% freeform (Appendix~Table~\ref{tab:format_validity}). Models that produce valid structured output lose just as much accuracy as those that fail to comply, so higher compliance does not translate into better reasoning.\label{sec:format_validity}

Taken together, the dominant cost is not imposed by the decoder but enters at the format-requesting prompt, before any decoder constraint is applied. Grammar-constrained decoding merely reinforces a constraint the model is already attempting to satisfy, bounding the ceiling for decoder-level fixes.

\subsection{Can the Format Tax Be Mitigated?}
\label{sec:two_turn}

If the format tax arises from reasoning and formatting in the same generation, separating these concerns should recover performance. Two natural approaches achieve this. Thinking allows the model to reason in a scratchpad before producing structured output---the format-requesting prompt is the same, but by the time the model generates its answer, it conditions on a full reasoning trace rather than the prompt alone. 2-Turn is a more explicit separation: the first call generates a freeform answer with no mention of formatting, and the second reformats it into the target structure. The cost of 2-Turn is two inference calls instead of one.


\begin{table}[t]
\centering
\small
\setlength{\tabcolsep}{4pt}
\resizebox{\textwidth}{!}{%
\begin{tabular}{@{}l rr@{\hspace{1em}}rr@{\hspace{1em}}rr@{\hspace{2em}}rr@{\hspace{1em}}rr@{\hspace{1em}}rr@{}}
\toprule
 & \multicolumn{6}{c}{No Thinking} & \multicolumn{6}{c}{With Thinking} \\
\cmidrule(lr){2-7} \cmidrule(lr){8-13}
 & \multicolumn{2}{c}{GPQA} & \multicolumn{2}{c}{Math} & \multicolumn{2}{c}{Zebra} & \multicolumn{2}{c}{GPQA} & \multicolumn{2}{c}{Math} & \multicolumn{2}{c}{Zebra} \\
\cmidrule(lr){2-3} \cmidrule(lr){4-5} \cmidrule(lr){6-7} \cmidrule(lr){8-9} \cmidrule(lr){10-11} \cmidrule(lr){12-13}
Model & 1T & 2T & 1T & 2T & 1T & 2T & 1T & 2T & 1T & 2T & 1T & 2T \\
\midrule
\qwenicon~qwen3-32b & \ccol{EFB5B5}-6 & \ccol{F3F9F3}+1 & \ccol{EDABAB}-7 & \ccol{F0F8F0}+2 & \ccol{EDAEAE}-7 & +1 & \ccol{FBEFEF}-1 & \ccol{B8DFB8}+8 & -0 & \ccol{F2F9F2}+2 & +1 & \ccol{ABD9AB}+10 \\
\qwenicon~qwen3-8b & \ccol{F3C9C9}-5 & \ccol{C7E6C7}+7 & \ccol{ECA7A7}-8 & \ccol{FCF2F2}-1 & \ccol{E58484}-11 & \ccol{F3F9F3}+1 & \ccol{EEF7EE}+2 & \ccol{ACDAAC}+10 & \ccol{F5FAF5}+1 & \ccol{EBF6EB}+2 & \ccol{F3F9F3}+1 & \ccol{B7DFB7}+9 \\
\allenaiicon~olmo3.1-32b & \ccol{F9E4E4}-2 & \ccol{BEE2BE}+8 & \ccol{F1C0C0}-5 & \ccol{E6F4E6}+3 & \ccol{E5F3E5}+3 & \ccol{F4FAF4}+1 & \ccol{F2F9F2}+2 & \ccol{AFDBAF}+10 & \ccol{FAEBEB}-2 & \ccol{E99898}-9 & -1 & \ccol{A8D8A8}+10 \\
\allenaiicon~olmo3-7b & \ccol{F2C3C3}-5 & \ccol{DDF0DD}+4 & \ccol{EFB3B3}-7 & \ccol{F6FBF6}+1 & \ccol{F6D5D5}-4 & \ccol{EDF7ED}+2 & \ccol{EBF6EB}+2 & \ccol{84C884}+15 & -0 & \ccol{DBEFDB}+4 & -1 & \ccol{8CCB8C}+14 \\
\hficon~smollm3-3b & \ccol{E5F3E5}+3 & \ccol{DFF0DF}+4 & \ccol{E79090}-10 & -0 & \ccol{F1BDBD}-6 & \ccol{F5FAF5}+1 & \ccol{E8F4E8}+3 & \ccol{C2E3C2}+7 & \ccol{F6D9D9}-3 & \ccol{F9E5E5}-2 & \ccol{FCF2F2}-1 & \ccol{ACDAAC}+10 \\
\nvidiaicon~nemotron3-nano & \ccol{EBA1A1}-8 & \ccol{EDAAAA}-7 & \ccol{EFB6B6}-6 & +1 & +1 & \ccol{D1EAD1}+5 & \ccol{DCEFDC}+4 & \ccol{AEDBAE}+10 & -0 & -0 & \ccol{BAE0BA}+8 & \ccol{B2DCB2}+9 \\
\bottomrule
\end{tabular}}%
\caption{Decoupling reverses the format tax.
Each cell shows the average accuracy change (\%) from the matched freeform baseline,
averaged across four formats and all prompt configurations.
1T = 1-Turn (single pass), 2T = 2-Turn (freeform then reformat).
Non-thinking columns use freeform as baseline;
thinking columns use freeform~+~thinking.}
\label{tab:decoupling-comparison}
\end{table}

Both approaches recover most of the format tax (Table~\ref{tab:decoupling-comparison}). 2-Turn significantly improves accuracy in 42 of 72 model--task--format comparisons ($p < 0.05$, average $+$6.8\,pp), while significantly worsening only 2. The improvement is consistent across tasks: MATH-500 (19 of 24 improved), GPQA (10 of 24), and ZebraLogic (13 of 24). On WritingBench, 2-Turn partially recovers quality (Table~\ref{tab:writingbench}). Thinking significantly improves 43 of 72 comparisons ($+$9.2\,pp average) but carries more risk: 11 comparisons (15\%) significantly worsen. The benefit is concentrated on MATH-500 (22 of 24 improved, 0 worsened) and GPQA (8 of 24 improved, 3 worsened); on ZebraLogic, results are mixed (13 improved, 8 worsened). Thinking does not transfer to writing: on WritingBench, it fails to close the gap and sometimes widens it.

2-Turn is the safer mitigation (3\% worsening rate), while thinking is more powerful but model- and task-dependent (15\% worsening rate). That both approaches work---despite differing in whether the format-requesting prompt is present---suggests the key factor is not the prompt text itself but what the model conditions on when it generates formatted output. We examine why thinking works despite sharing the same prompt in Section~\ref{sec:thinking_mechanism}.

\section{Discussion}
\label{sec:discussion}

\begin{wrapfigure}{r}{0.5\columnwidth}
\vspace{-1.5em}
\centering
\includegraphics[width=0.48\columnwidth]{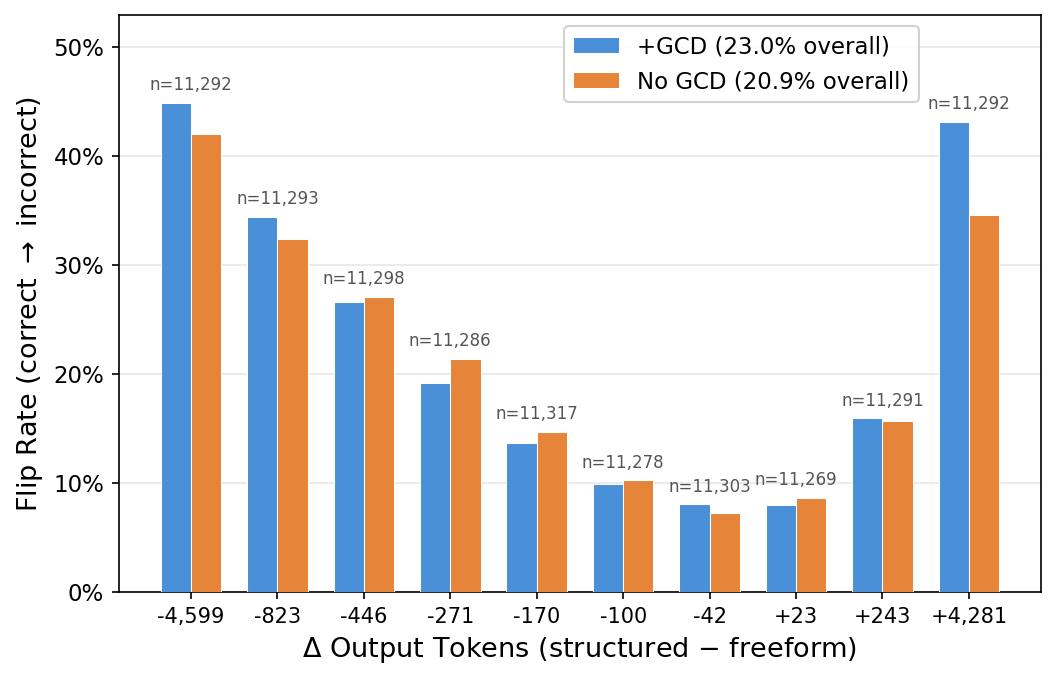}
\vspace{-0.5em}
\caption{Per-question flip rate (freeform-correct $\to$ structured-incorrect) by token delta decile. The U-shaped pattern is nearly identical with and without GCD.}
\label{fig:flip_rate}
\vspace{-1em}
\end{wrapfigure}
\paragraph{Why does thinking recover the format tax?}
\label{sec:thinking_mechanism}
Thinking nearly eliminates the format tax on reasoning tasks (Table~\ref{tab:decoupling-comparison}), but the model still conditions on the same format-requesting prompt.
A per-question view suggests that the tax concentrates where structured output most compresses the model's natural output length.\label{sec:token_compression} Among roughly 110{,}000 paired observations, 71\% of structured outputs are shorter than their freeform counterparts (median $\Delta = -162$ tokens). Flip rates follow a U-shaped curve: lowest near zero token delta and exceeding 40\% at the strongest compression or expansion extremes (Figure~\ref{fig:flip_rate}). The same pattern appears with and without GCD and holds across all six models individually (Appendix~Figure~\ref{fig:token_flip_rate_by_model}), indicating that answer flips arise from the model's response to the format request rather than decoder-level masking.

Thinking recovers the tax by separating reasoning from formatting within a single generation. Thinking token counts remain strongly aligned between freeform and structured output ($r = 0.72$ across 28{,}000+ uncapped pairs; Appendix~Figure~\ref{fig:thinking_token_correlation}), and format-related content accounts for less than 3\% of thinking sentences (Appendix~Table~\ref{tab:thinking_format_leakage}). Format prompts compress or redirect visible reasoning, causing answer flips, but thinking creates a temporally separate reasoning phase before the model commits to the requested structure.

\paragraph{The closed-weight gap.}
\label{sec:closed_weight_gap}
Newer closed-weight models point to a different regime. On our reasoning benchmarks, \model{claude-haiku-4.5}, \model{grok-4.1-fast}, and \model{gpt-5.4-nano} show little to no format tax, whereas every open-weight model we evaluate still degrades under structured output. Because these systems are closed-weight, we cannot tell whether this robustness reflects deliberate structured-output fine-tuning, broader post-training, or an emergent capability of scale. What the comparison does establish is that the tax is not inherent to structured generation itself: some models already resist it. That raises a practical question for the open-weight community: should robustness to format constraints become an explicit training objective?

\section{Related Work}
\label{sec:related_work}

\citet{tam2024letspeakfreelystudy} documented that format constraints degrade LLM accuracy. The empirical response has targeted the decoder: grammar-aligned decoding methods \citep{gad, cars, gad_mcmc, awrs, genlm_smc} and iterative backtracking \citep{ugare2025itergeniterativesemanticawarestructured} attempt to correct sampling bias introduced by token masking. Our decomposition bounds the ceiling for these decoder-level fixes: the gap between constrained and unconstrained decoding under the same format-requesting prompt is small, indicating that the primary cause lies upstream.

On the theoretical side, \citet{crane} prove that finite output grammars without reasoning tokens reduce model expressivity to TC$^0$, providing formal justification for decoupling reasoning from formatting---a principle consistent with our 2-Turn results. \citet{dccd} formalize constrained decoding as repeated distribution projections and propose draft-conditioned constrained decoding, which conditions on a freeform draft to recover accuracy while maintaining structural guarantees. \citet{nguyen2025thinking} achieve similar recovery in a single pass by allowing free reasoning before a trigger activates constrained decoding.

Not all evidence points toward degradation. In logical parsing, GCD can improve accuracy by preventing malformed outputs \citep{raspanti2025grammar}, suggesting the tax depends on whether format compliance and task correctness are aligned or in competition. GCD speed optimization \citep{disc, dong2025xgrammarflexibleefficientstructured} and format validity benchmarking \citep{geng_jsonschemabench} address complementary concerns; our work asks whether enforcing validity comes at a cost to capability.

\section{Limitations}
\label{sec:limitations}

Our findings apply specifically to \emph{format} constraints---grammars that enforce output structure for presentation purposes. Grammar-constrained decoding has other applications (code synthesis, function calling, test generation) where constraints encode correctness requirements rather than presentation preferences; our results should not be extrapolated to those settings.

Our open-weight models span 3B--32B parameters; larger open-weight models may behave differently. Task selection emphasized reasoning-heavy benchmarks where thinking competes with formatting for generation capacity; simpler extraction tasks may show smaller effects. We tested four formats with relatively simple schemas; more complex structures or untested formats (YAML, TOML, protocol buffers) could produce different results. The closed-weight analysis covers only four API models from three providers---too few to draw general conclusions about whether format-tax resistance is widespread among frontier models or specific to these systems.

The 2-Turn strategy that recovers most lost performance introduces practical costs: additional API calls, increased latency, and roughly doubled token consumption. For latency-sensitive applications, this overhead may be prohibitive.

Our controlled setting likely underestimates the format tax in practice. Real deployments contend with schema errors, ambiguous specifications, and malformed examples---conditions that would compound the degradation we identified.

\section{Conclusion}
\label{sec:conclusion}

Requiring structured output degrades reasoning and writing performance across every open-weight model we tested, while recent closed-weight models resist the effect. The degradation enters at the prompt, not the decoder: grammar-constrained decoding adds little beyond what the format-requesting prompt already imposes.

The most effective mitigation is to decouple reasoning from formatting---either by generating freeform first and reformatting in a second pass, or by enabling extended thinking within a single generation. Both recover most of the lost accuracy; varying prompt content alone does not.

The tax is not a fundamental limitation. Some models already resist it, and decoupled generation reverses it without retraining, suggesting that the underlying reasoning capability remains intact. What remains unclear is how resistant models achieved this robustness and whether open-weight models can close the gap through training, scale, or both.

\section*{Ethics Statement}
This work evaluates existing language models on publicly available benchmarks and does not involve human subjects or private data. Our finding that structured output requirements degrade reasoning could inform deployment decisions; we see no direct dual-use risk from documenting this effect.

\section*{Reproducibility Statement}
All experiments use publicly available models and benchmarks. Open-weight models run via vLLM; API models use provider-native structured output endpoints. The full experimental matrix (Table~\ref{tab:strategies}), model list (Section~\ref{sec:experimental_design}), and hyperparameters (Appendix~\ref{sec:appendix_details}) are reported. MATH-500 scoring uses an LLM judge whose model and settings are documented in Appendix~\ref{sec:math500_scoring}. Code and reproduction instructions are available at \url{https://github.com/ivnle/the-format-tax}.

\section*{LLM Disclosure}
We used \model{gpt-5.4-nano} as a math-equivalence judge for MATH-500 scoring (Appendix~\ref{sec:math500_scoring}) and \model{gpt-5.2} as a writing-quality judge for WritingBench. LLMs were used to assist with drafting and revising paper text. No LLMs were used to originate research ideas or generate experimental results.

\bibliography{colm2026}

\begin{thebibliography}{21}
\providecommand{\natexlab}[1]{#1}
\providecommand{\url}[1]{\texttt{#1}}
\expandafter\ifx\csname urlstyle\endcsname\relax
  \providecommand{\doi}[1]{doi: #1}\else
  \providecommand{\doi}{doi: \begingroup \urlstyle{rm}\Url}\fi

\bibitem[Banerjee et~al.(2025)Banerjee, Suresh, Ugare, Misailovic, and
  Singh]{crane}
Debangshu Banerjee, Tarun Suresh, Shubham Ugare, Sasa Misailovic, and Gagandeep
  Singh.
\newblock Crane: Reasoning with constrained llm generation, 2025.
\newblock URL \url{https://arxiv.org/abs/2502.09061}.

\bibitem[Dong et~al.(2025)Dong, Ruan, Cai, Lai, Xu, Zhao, and
  Chen]{dong2025xgrammarflexibleefficientstructured}
Yixin Dong, Charlie~F. Ruan, Yaxing Cai, Ruihang Lai, Ziyi Xu, Yilong Zhao, and
  Tianqi Chen.
\newblock Xgrammar: Flexible and efficient structured generation engine for
  large language models, 2025.
\newblock URL \url{https://arxiv.org/abs/2411.15100}.

\bibitem[Geng et~al.(2024)Geng, Josifoski, Peyrard, and West]{gcd}
Saibo Geng, Martin Josifoski, Maxime Peyrard, and Robert West.
\newblock Grammar-constrained decoding for structured nlp tasks without
  finetuning, 2024.
\newblock URL \url{https://arxiv.org/abs/2305.13971}.

\bibitem[Geng et~al.(2025)Geng, Cooper, Moskal, Jenkins, Berman, Ranchin, West,
  Horvitz, and Nori]{geng_jsonschemabench}
Saibo Geng, Hudson Cooper, Michał Moskal, Samuel Jenkins, Julian Berman,
  Nathan Ranchin, Robert West, Eric Horvitz, and Harsha Nori.
\newblock Jsonschemabench: A rigorous benchmark of structured outputs for
  language models, 2025.
\newblock URL \url{https://arxiv.org/abs/2501.10868}.

\bibitem[Gonzalez et~al.(2025)Gonzalez, Vaidya, Park, Ji, Berg-Kirkpatrick, and
  D'Antoni]{gad_mcmc}
Emmanuel~Anaya Gonzalez, Sairam Vaidya, Kanghee Park, Ruyi Ji, Taylor
  Berg-Kirkpatrick, and Loris D'Antoni.
\newblock Constrained sampling for language models should be easy: An mcmc
  perspective, 2025.
\newblock URL \url{https://arxiv.org/abs/2506.05754}.

\bibitem[Hendrycks et~al.(2021)Hendrycks, Burns, Kadavath, Arora, Basart, Tang,
  Song, and Steinhardt]{hendrycks2021math}
Dan Hendrycks, Collin Burns, Saurav Kadavath, Akul Arora, Steven Basart, Eric
  Tang, Dawn Song, and Jacob Steinhardt.
\newblock Measuring mathematical problem solving with the math dataset, 2021.
\newblock URL \url{https://arxiv.org/abs/2103.03874}.

\bibitem[Lin et~al.(2025)Lin, Gupta, Jain, Ananthakrishnan, Chandu, Brahman,
  and Ren]{lin2025zebralogic}
Bill~Yuchen Lin, Aditya Gupta, Naman Jain, Shankar Ananthakrishnan,
  Khyathi~Raghavi Chandu, Faeze Brahman, and Xiang Ren.
\newblock Zebralogic: On the scaling limits of llms for logical reasoning,
  2025.
\newblock URL \url{https://arxiv.org/abs/2502.01100}.

\bibitem[Lipkin et~al.(2025)Lipkin, LeBrun, Vigly, Loula, MacIver, Du, Eisner,
  Cotterell, Mansinghka, O'Donnell, Lew, and Vieira]{awrs}
Benjamin Lipkin, Benjamin LeBrun, Jacob~Hoover Vigly, João Loula, David~R.
  MacIver, Li~Du, Jason Eisner, Ryan Cotterell, Vikash Mansinghka, Timothy~J.
  O'Donnell, Alexander~K. Lew, and Tim Vieira.
\newblock Fast controlled generation from language models with adaptive
  weighted rejection sampling, 2025.
\newblock URL \url{https://arxiv.org/abs/2504.05410}.

\bibitem[Loula et~al.(2025)Loula, LeBrun, Du, Lipkin, Pasti, Grand, Liu, Emara,
  Freedman, Eisner, Cotterell, Mansinghka, Lew, Vieira, and
  O'Donnell]{genlm_smc}
João Loula, Benjamin LeBrun, Li~Du, Ben Lipkin, Clemente Pasti, Gabriel Grand,
  Tianyu Liu, Yahya Emara, Marjorie Freedman, Jason Eisner, Ryan Cotterell,
  Vikash Mansinghka, Alexander~K. Lew, Tim Vieira, and Timothy~J. O'Donnell.
\newblock Syntactic and semantic control of large language models via
  sequential monte carlo, 2025.
\newblock URL \url{https://arxiv.org/abs/2504.13139}.

\bibitem[Nguyen et~al.(2025)Nguyen, Silva, Zumot, Tupikina, Aghasaryan, and
  Alam]{nguyen2025thinking}
Ngoc Trinh~Hung Nguyen, Alonso Silva, Laith Zumot, Liubov Tupikina, Armen
  Aghasaryan, and Mehwish Alam.
\newblock Thinking before constraining: A unified decoding framework for large
  language models, 2025.
\newblock URL \url{https://arxiv.org/abs/2601.07525}.

\bibitem[Park et~al.(2024)Park, Wang, Berg-Kirkpatrick, Polikarpova, and
  D'Antoni]{gad}
Kanghee Park, Jiayu Wang, Taylor Berg-Kirkpatrick, Nadia Polikarpova, and Loris
  D'Antoni.
\newblock Grammar-aligned decoding, 2024.
\newblock URL \url{https://arxiv.org/abs/2405.21047}.

\bibitem[Park et~al.(2025)Park, Zhou, and
  D'Antoni]{park2025flexibleefficientgrammarconstraineddecoding}
Kanghee Park, Timothy Zhou, and Loris D'Antoni.
\newblock Flexible and efficient grammar-constrained decoding, 2025.
\newblock URL \url{https://arxiv.org/abs/2502.05111}.

\bibitem[Parys et~al.(2025)Parys, Vaidya, Berg-Kirkpatrick, and D'Antoni]{cars}
Pawe{\l} Parys, Sairam Vaidya, Taylor Berg-Kirkpatrick, and Loris D'Antoni.
\newblock Constrained adaptive rejection sampling, 2025.
\newblock URL \url{https://arxiv.org/abs/2510.01902}.

\bibitem[Raspanti et~al.(2025)Raspanti, Ozcelebi, and
  Holenderski]{raspanti2025grammar}
Federico Raspanti, Tanir Ozcelebi, and Mike Holenderski.
\newblock Grammar-constrained decoding makes large language models better
  logical parsers.
\newblock In \emph{Proceedings of the 63rd Annual Meeting of the Association
  for Computational Linguistics (Volume 6: Industry Track)}, pp.\  485--499,
  jul 2025.
\newblock \doi{10.18653/v1/2025.acl-industry.34}.
\newblock URL \url{https://aclanthology.org/2025.acl-industry.34/}.

\bibitem[Reddy et~al.(2026)Reddy, Walker, Ide, and Bedi]{dccd}
Avinash Reddy, Thayne~T. Walker, James~S. Ide, and Amrit~Singh Bedi.
\newblock Draft-conditioned constrained decoding for structured generation in
  llms, 2026.
\newblock URL \url{https://arxiv.org/abs/2603.03305}.

\bibitem[Rein et~al.(2023)Rein, Hou, Stickland, Petty, Pang, Dirani, Michael,
  and Bowman]{rein2023gpqa}
David Rein, Betty~Li Hou, Asa~Cooper Stickland, Jackson Petty, Richard~Yuanzhe
  Pang, Julien Dirani, Julian Michael, and Samuel~R. Bowman.
\newblock Gpqa: A graduate-level google-proof q\&a benchmark, 2023.
\newblock URL \url{https://arxiv.org/abs/2311.12022}.

\bibitem[Tam et~al.(2024)Tam, Wu, Tsai, Lin, yi~Lee, and
  Chen]{tam2024letspeakfreelystudy}
Zhi~Rui Tam, Cheng-Kuang Wu, Yi-Lin Tsai, Chieh-Yen Lin, Hung yi~Lee, and
  Yun-Nung Chen.
\newblock Let me speak freely? a study on the impact of format restrictions on
  performance of large language models, 2024.
\newblock URL \url{https://arxiv.org/abs/2408.02442}.

\bibitem[Ugare et~al.(2025)Ugare, Gumaste, Suresh, Singh, and
  Misailovic]{ugare2025itergeniterativesemanticawarestructured}
Shubham Ugare, Rohan Gumaste, Tarun Suresh, Gagandeep Singh, and Sasa
  Misailovic.
\newblock Itergen: Iterative semantic-aware structured llm generation with
  backtracking, 2025.
\newblock URL \url{https://arxiv.org/abs/2410.07295}.

\bibitem[Willard \& Louf(2023)Willard and
  Louf]{willard2023efficientguidedgenerationlarge}
Brandon~T. Willard and Rémi Louf.
\newblock Efficient guided generation for large language models, 2023.
\newblock URL \url{https://arxiv.org/abs/2307.09702}.

\bibitem[Wu et~al.(2025)Wu, Mei, Yan, Li, Lai, Ren, Wang, Zhang, Wu, Jin, and
  Huang]{wu2025writingbench}
Yuning Wu, Jiahao Mei, Ming Yan, Chenliang Li, Shaopeng Lai, Yuran Ren, Zijia
  Wang, Ji~Zhang, Mengyue Wu, Qin Jin, and Fei Huang.
\newblock Writingbench: A comprehensive benchmark for generative writing, 2025.
\newblock URL \url{https://arxiv.org/abs/2503.05244}.

\bibitem[Ye et~al.(2025)Ye, Jain, You, Suresh, Lin, Zou, and Yu]{disc}
Haotian Ye, Himanshu Jain, Chong You, Ananda~Theertha Suresh, Haowei Lin, James
  Zou, and Felix Yu.
\newblock Efficient and asymptotically unbiased constrained decoding for large
  language models, 2025.
\newblock URL \url{https://arxiv.org/abs/2504.09135}.

\end{thebibliography}
\bibliographystyle{colm2026_conference}

\appendix


\section{Format Constraint Definitions}
\label{sec:format-constraints}

We define format constraints using JSON Schema (for JSON) and Lark grammars
(for XML, Markdown, and \LaTeX).  Each constraint defines a subset of the
format tailored to the task's output structure---specifically the fields the
model must populate and their nesting.  Below we show the MATH-500 constraints
and representative valid outputs produced by Nemotron-Nano-8B under each
format.

\noindent
\begin{minipage}[t]{0.48\textwidth}
\begin{tcolorbox}[title={\textbf{JSON} --- JSON Schema constraint},
  colback=blue!3, colframe=blue!40!black, fonttitle=\sffamily\small, boxsep=2pt, top=2pt, bottom=2pt]
\begin{lstlisting}[basicstyle=\ttfamily\tiny, breaklines=true,
  autogobble, language={}, xleftmargin=0pt]
{
  "properties": {
    "reasoning": {
      "description": "Step-by-step reasoning to solve the problem",
      "type": "string"
    },
    "answer": {
      "description": "Final answer to the problem",
      "type": "string"
    }
  },
  "required": ["reasoning", "answer"],
  "type": "object"
}
\end{lstlisting}

\tcblower
\textbf{Example output:}
\begin{lstlisting}[basicstyle=\ttfamily\tiny, breaklines=true,
  autogobble, language={}]
{
  "reasoning": "To convert rectangular (0, 3) to polar:
    r = sqrt(0^2 + 3^2) = 3.
    theta = atan2(3, 0) = pi/2. ...",
  "answer": "(3, pi/2)"
}
\end{lstlisting}
\end{tcolorbox}
\end{minipage}\hfill
\begin{minipage}[t]{0.48\textwidth}
\begin{tcolorbox}[title={\textbf{XML} --- Lark grammar constraint},
  colback=green!3, colframe=green!40!black, fonttitle=\sffamily\small, boxsep=2pt, top=2pt, bottom=2pt]
\begin{lstlisting}[basicstyle=\ttfamily\tiny, breaklines=true,
  autogobble, language={}]
start: "<response>" _WS? reasoning _WS? answer
       _WS? "</response>" _WS?

reasoning: "<reasoning>" text "</reasoning>"
answer:    "<answer>" text "</answer>"

text: /[^<>]+/
_WS:  /[ \t\n]+/
\end{lstlisting}

\tcblower
\textbf{Example output:}
\begin{lstlisting}[basicstyle=\ttfamily\tiny, breaklines=true,
  autogobble, language=XML]
<response>
<reasoning>
To convert rectangular (0,3) to polar:
  r = sqrt(0^2 + 3^2) = 3.
  theta = atan2(3,0) = pi/2. ...</reasoning>
<answer>(3, pi/2)</answer>
</response>
\end{lstlisting}
\end{tcolorbox}
\end{minipage}

\smallskip

\noindent
\begin{minipage}[t]{0.48\textwidth}
\begin{tcolorbox}[title={\textbf{Markdown} --- Lark grammar constraint},
  colback=orange!3, colframe=orange!40!black, fonttitle=\sffamily\small, boxsep=2pt, top=2pt, bottom=2pt]
\begin{lstlisting}[basicstyle=\ttfamily\tiny, breaklines=true,
  autogobble, language={}]
start: reasoning_section _WS? answer_section _WS?

reasoning_section: "## Reasoning" _WS? text
answer_section:    "## Answer" _WS? text

text: line+
line: /[^#\n][^\n]*\n?/   // non-# start
    | /#[^#\n][^\n]*\n?/  // single # (not ##)
    | /#\n/                // bare #
    | /\n/                 // blank line

_WS: /[ \t\n]+/
\end{lstlisting}

\tcblower
\textbf{Example output:}
\begin{lstlisting}[basicstyle=\ttfamily\tiny, breaklines=true,
  autogobble, language={}]
## Reasoning

To convert (0, 3) to polar coordinates:
r = sqrt(0^2 + 3^2) = sqrt(9) = 3.
Since x = 0 and y > 0, theta = pi/2. ...

## Answer

(3, pi/2)
\end{lstlisting}
\end{tcolorbox}
\end{minipage}\hfill
\begin{minipage}[t]{0.48\textwidth}
\begin{tcolorbox}[title={\textbf{\LaTeX} --- Lark grammar constraint (abbreviated)},
  colback=red!3, colframe=red!40!black, fonttitle=\sffamily\small, boxsep=2pt, top=2pt, bottom=2pt]
\begin{lstlisting}[basicstyle=\ttfamily\tiny, breaklines=true,
  autogobble, language={}, escapechar=@]
start: _WS? preamble _WS? document _WS?

preamble: documentclass preamble_content*
documentclass: "\\documentclass" class_options? "{" CLASS_NAME "}"
  // ... (usepackage, generic_command rules omitted)

document: "\\begin{document}" _WS? body _WS?
          "\\end{document}"
body: _WS? response _WS?
response: "\\begin{response}" _WS? reasoning _WS?
          answer _WS? "\\end{response}"

reasoning: "\\reasoning{" text "}"
answer:    "\\answer{" text "}"

text: text_element+
text_element: text_char | brace_group
  // ... (brace nesting and escape rules omitted)

_WS: /[ \t\n]+/
\end{lstlisting}

\tcblower
\textbf{Example output:}
\begin{lstlisting}[basicstyle=\ttfamily\tiny, breaklines=true,
  autogobble, language={[LaTeX]TeX}, texcl=false,
  morekeywords={reasoning,answer,response}]
\documentclass{article}
\usepackage{amsmath}
\begin{document}
\begin{response}
\reasoning{To convert (0, 3) to polar:
  $r = \sqrt{0^2 + 3^2} = 3$.
  Since $x = 0$ and $y > 0$,
  $\theta = \frac{\pi}{2}$. ...}
\answer{$(3,\,\frac{\pi}{2})$}
\end{response}
\end{document}
\end{lstlisting}
\end{tcolorbox}
\end{minipage}

\FloatBarrier

\section{Cases Where Structured Output Improves Accuracy}
\label{sec:positive_cells}

A small number of cells in Tables~\ref{tab:format_tax_universal_ow} and \ref{tab:format_tax_universal_api} show accuracy improvements under structured output (green entries). These trace to four model-specific mechanisms rather than a systematic benefit of formatting.

\paragraph{Floor effects.}
\model{smollm3-3b}'s GPQA gains arise because the model scores below chance in freeform. Constraining output to valid answer tokens narrows the search space without requiring better reasoning.

\paragraph{Degenerate freeform outputs.}
\model{olmo3.1-32b}'s positive ZebraLogic cells trace to freeform output collapse: 16\% of responses degenerate to low-effort guesses with no reasoning. Structured format prevents this by requiring a populated reasoning field. Excluding degenerate outputs, three of four green cells flip to neutral or negative.

\paragraph{Structured output as reasoning scaffolding.}
\model{gpt-5.4-nano} and \model{grok-4.1-fast} show positive ZebraLogic deltas (+4--13\% across formats). Paired analysis confirms these are not scoring artifacts: in both models, every structured-win instance contains a correct answer. Manual categorization of 100 random structured-win pairs per model shows that freeform failures are predominantly genuine reasoning errors---the model arrives at a wrong entity, not merely a wrong letter. The structured template appears to scaffold more reliable constraint propagation on multi-step logic puzzles. These gains are specific to ZebraLogic; neither model shows positive deltas on GPQA or MATH-500.

\paragraph{Format-induced strategy redirection.}
\model{nemotron3-nano}'s ZebraLogic gains arise from two mechanisms. For JSON, freeform responses open with a markdown table that the model fills before fully reasoning through constraints, causing premature commitment and cascading errors. JSON forces sequential reasoning in a dedicated field, redirecting the model to a more effective strategy. For XML, the green cell is driven by a single prompt configuration where freeform outputs collapse to ultra-short answers (median 160 tokens). XML's reasoning tag forces thoroughness (median 2{,}000+ tokens), acting as a guardrail against shortcutting.

\FloatBarrier

\section{Non-JSON GCD Collapse on \model{gpt-5.4-nano}}
\label{sec:api_gcd_collapse}

\model{gpt-5.4-nano} shows severe accuracy degradation under grammar-constrained decoding for non-JSON formats (XML: $-$59\,pp, LaTeX: $-$28\,pp in Table~\ref{tab:format_tax_universal_api}), while JSON GCD and all no-GCD configurations remain near baseline. Comparing the same formats with and without GCD confirms the grammar constraint is the cause: token counts drop from 130--200 to 50--90 as the model generates placeholder answers---e.g., \texttt{<answer>Proceed to solve by algebraic constraints.</answer>}---instead of performing the calculation.

This collapse is model-specific. The previous-generation \model{gpt-5-nano} achieves 86.7\% on the identical XML+GCD prompt (vs.\ 29.5\% for \model{gpt-5.4-nano}). The likely explanation is that \model{gpt-5.4-nano}'s structured output training was optimized for JSON and did not generalize to non-JSON grammars.

\FloatBarrier

\section{Thinking and Temporal Separation}
\label{sec:appendix_thinking}

\begin{figure}[t]
\centering
\includegraphics[width=0.7\columnwidth]{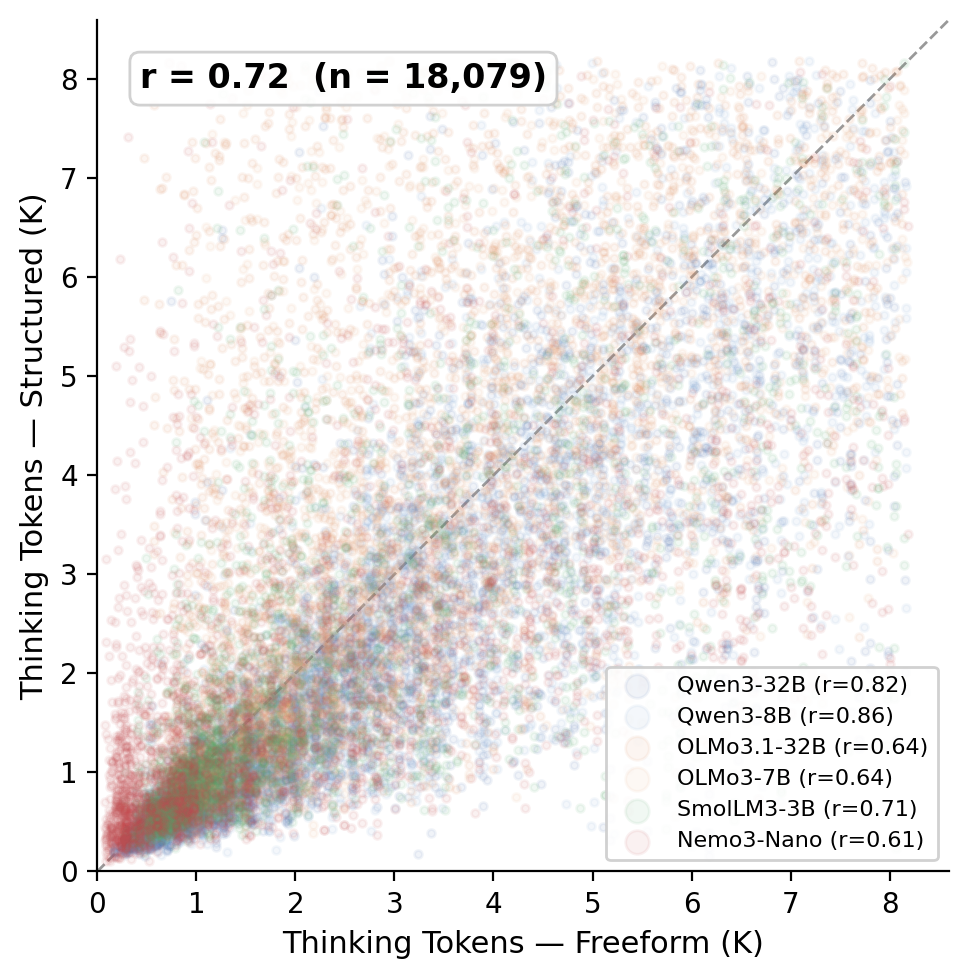}
\caption{Per-question thinking tokens under freeform vs.\ structured output. Each point is one question; pairs where either side hits the 8{,}192-token ceiling are excluded. The strong correlation ($r = 0.72$, $n = 28{,}000$+) indicates that thinking effort is driven by question difficulty, not format constraints.}
\label{fig:thinking_token_correlation}
\end{figure}

If format constraints diverted thinking effort, we would expect models to spend more thinking tokens under structured output. Figure~\ref{fig:thinking_token_correlation} shows this is not the case: thinking effort tracks question difficulty, not format. Table~\ref{tab:thinking_format_leakage} further checks whether format markup leaks into thinking blocks.

\begin{table}[t]
\centering
\small
\begin{tabular}{@{}l ccc@{}}
\toprule
 & Structured & Freeform & Net \\
 & (\%) & (\%) & (pp) \\
\midrule
qwen3-32b & 0.7 & 0.7 & +0.0 \\
qwen3-8b & 0.7 & 0.8 & -0.0 \\
olmo3.1-32b & 3.0 & 0.8 & +2.2 \\
olmo3-7b & 2.0 & 0.7 & +1.3 \\
smollm3-3b & 0.6 & 0.6 & -0.0 \\
nemotron3-nano & 3.0 & 0.6 & +2.4 \\
\bottomrule
\end{tabular}
\caption{Format leakage into thinking blocks.
Fraction of sentences in \texttt{\textless think\textgreater} blocks that mention
format-related keywords (JSON, XML, format, structure, etc.).
Freeform: same metric as a false-positive control (these keywords appear
incidentally in math/logic reasoning).
Net: structured minus freeform.
Despite this, thinking nearly eliminates the format tax (Table~\ref{tab:decoupling-comparison}).}
\label{tab:thinking_format_leakage}
\end{table}

\FloatBarrier

\section{Format Validity}
\label{sec:appendix_format_validity}

Table~\ref{tab:format_validity} reports format validity rates alongside accuracy. Grammar-constrained decoding guarantees valid output by construction; without it, most models still produce valid output at high rates, confirming that accuracy differences are not driven by parse failures.

\begin{table}[t]
\centering
\small
\begin{tabular}{@{}l cc cc@{}}
\toprule
 & \multicolumn{2}{c}{No GCD} & \multicolumn{2}{c}{+GCD} \\
\cmidrule(lr){2-3} \cmidrule(lr){4-5}
 & Valid & Acc & Valid & Acc \\
\midrule
JSON & \textbf{65.0} & 56.4 & 96.2 & 54.3 \\
XML & 60.7 & 57.4 & \textbf{97.0} & 54.5 \\
Markdown & 46.5 & \textbf{58.6} & 94.3 & \textbf{57.1} \\
LaTeX & 50.5 & 56.8 & 81.4 & 57.0 \\
\midrule
qwen3-32b & 76.4 & 61.3 & \textbf{98.9} & 59.1 \\
qwen3-8b & \textbf{77.6} & 55.9 & 95.6 & 53.8 \\
olmo3.1-32b & 64.6 & \textbf{71.8} & 90.5 & \textbf{70.4} \\
olmo3-7b & 39.3 & 64.2 & 91.2 & 61.0 \\
smollm3-3b & 36.1 & 39.5 & 86.1 & 38.1 \\
nemotron3-nano & 40.1 & 51.3 & 91.0 & 51.9 \\
\midrule
All & 55.7 & 57.3 & 92.2 & 55.7 \\
\bottomrule
\end{tabular}
\caption{Format validity and lenient accuracy (correct answer regardless of format compliance) in \% by decoding method.
+GCD dramatically improves format compliance (55.7\% $\to$ 92.2\%)
but does not improve reasoning accuracy, which remains
5--7\,pp below the freeform baseline (61.5\%)
regardless of compliance.
All values averaged across 1-turn, thinking-off experiments
on three reasoning benchmarks.}
\label{tab:format_validity}
\end{table}

\FloatBarrier

\section{Per-Model Flip Rate}
\label{sec:appendix_flip_rate}

\begin{figure}[H]
\centering
\includegraphics[width=\textwidth]{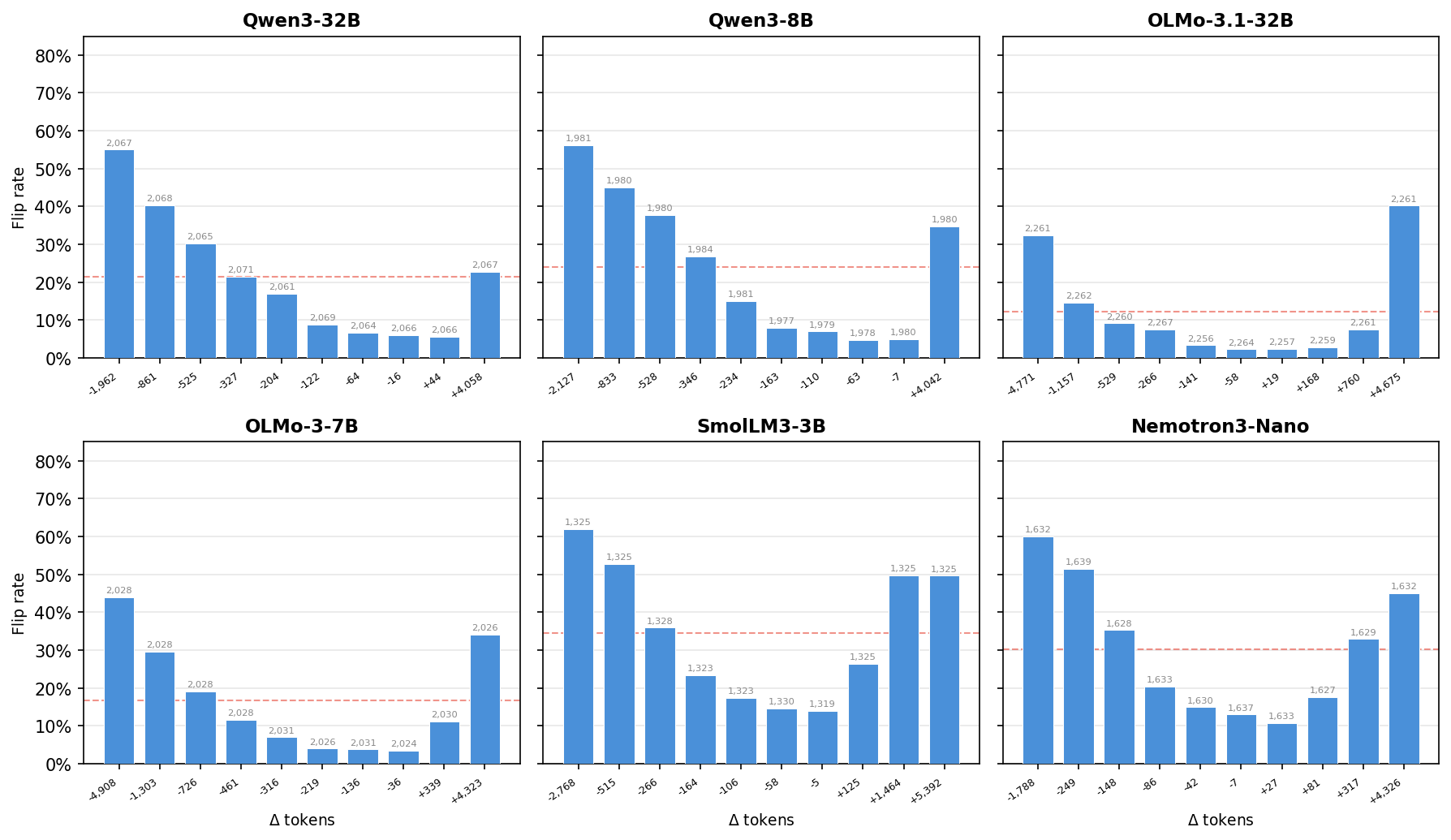}
\caption{Per-model flip rate by token delta decile. The U-shaped pattern from Figure~\ref{fig:flip_rate} holds across all six models: questions where structured output most compresses or expands token count relative to freeform show the highest flip rates.}
\label{fig:token_flip_rate_by_model}
\end{figure}

Figure~\ref{fig:token_flip_rate_by_model} disaggregates the flip rate analysis from Figure~\ref{fig:flip_rate} by model. The U-shaped pattern holds across all models, confirming that the relationship between token delta and accuracy loss is not driven by any single model.

\FloatBarrier

\section{MATH-500 Scoring}
\label{sec:math500_scoring}

GPQA and ZebraLogic are multiple-choice: deterministic letter extraction is reliable across all output formats. MATH-500 is open-ended, and its answers---integers, fractions, symbolic expressions---appear differently depending on format markup. We initially used the same deterministic pipeline for all reasoning tasks. On MATH-500, extraction failure rates were low for freeform but high for structured formats, inflating the measured tax. Manual inspection confirmed these were parsing failures, not reasoning errors: models produced correct answers that the extractor could not handle. Format-specific extractors would have introduced format-dependent scoring leniency, defeating the purpose of a controlled comparison.

We replaced this with a fixed LLM judge: \model{gpt-5.4-nano} (reasoning\_effort=low, temperature 0) receives the model's response and the gold answer and determines whether they are mathematically equivalent. The switch reduced the average MATH-500 tax from $-$10.1\,pp to $-$6.6\,pp, confirming that deterministic extraction was overcounting the tax. \model{gpt-5.4-nano} also appears as an evaluated model; we use it as judge because comparing two expressions is distinct from the evaluated task of multi-step problem solving. Because the judge reduces the measured tax, our reported numbers are conservative relative to deterministic scoring.

Table~\ref{tab:math500-scoring-agreement} quantifies disagreement rates across formats: freeform agreement is near-perfect (96.6\%), while structured formats show higher divergence concentrated in deterministic failures that the judge recovers. Reverse disagreements (deterministic pass, judge fail) remain near zero across all formats. Table~\ref{tab:math500-scoring-examples} shows representative examples of each failure mode.


\begin{table*}[t]
\centering
\footnotesize
\setlength{\tabcolsep}{4pt}
\resizebox{\textwidth}{!}{%
\begin{tabular}{@{}l r r r r r r@{}}
\toprule
Format & $N$ judged & Both Correct & Both Incorrect & Det.\ \xmark\ / Judge \cmark & Det.\ \cmark\ / Judge \xmark & Agreement \\
\midrule
Freeform & 4970 & 3986 (80.2\%) & 814 (16.4\%) & 164 (3.3\%) & 6 (0.1\%) & 4800 (96.6\%) \\
JSON & 4981 & 3700 (74.3\%) & 1056 (21.2\%) & 220 (4.4\%) & 5 (0.1\%) & 4756 (95.5\%) \\
XML & 4973 & 3713 (74.7\%) & 1012 (20.3\%) & 242 (4.9\%) & 6 (0.1\%) & 4725 (95.0\%) \\
Markdown & 4965 & 3565 (71.8\%) & 872 (17.6\%) & 522 (10.5\%) & 6 (0.1\%) & 4437 (89.4\%) \\
LaTeX & 4976 & 3672 (73.8\%) & 1012 (20.3\%) & 288 (5.8\%) & 4 (0.1\%) & 4684 (94.1\%) \\
\bottomrule
\end{tabular}}%
\caption{Agreement between the deterministic MATH-500 scorer and the LLM judge on the canonical ten-model, no-GCD slice. Counts and percentages are computed over judged rows only. Divergence is concentrated in deterministic failures that the judge recovers, while reverse disagreements remain rare.}
\label{tab:math500-scoring-agreement}
\end{table*}


\begin{table*}[t]
\centering
\footnotesize
\setlength{\tabcolsep}{4pt}
\begin{tabularx}{\textwidth}{@{}l >{\raggedright\arraybackslash}X >{\raggedright\arraybackslash}p{0.14\textwidth} c c@{}}
\toprule
Format & Model Response & Gold & Det. & Judge \\
\midrule
JSON & 10\% & $10$ & \xmark & \cmark \\
Markdown & $x = 3,\; x = 5,\; x = 7$ & $3, 5, 7$ & \xmark & \cmark \\
JSON & quotient $= x^5 - x^4 + x^3 - x^2 + x - 1$; remainder $= {-}2$ & $x^5 - x^4 + \cdots + x - 1$ & \xmark & \cmark \\
LaTeX & $11\sqrt{2}$ & $11\sqrt2$ & \xmark & \cmark \\
XML & $864$ & $864\ \text{in}^2$ & \xmark & \cmark \\
Markdown & Final answer: $1$ (also mentions $-2$ earlier) & $-2$ & \cmark & \xmark \\
Freeform & $6$ & $6$ & \cmark & \cmark \\
Freeform & $9$ & $13$ & \xmark & \xmark \\
\bottomrule
\end{tabularx}
\caption{Reviewed real MATH-500 rows illustrating the appendix audit. The disagreements are concentrated in extraction, parse, and normalization failures of the deterministic scorer; the reverse-disagreement example shows the judge rejecting a response whose final answer is actually wrong.}
\label{tab:math500-scoring-examples}
\end{table*}

Our main claims also hold on GPQA and ZebraLogic, which use no LLM judge.

\FloatBarrier

\section{Full Experimental Results}
\label{sec:appendix_full_results}

\scriptsize
\setlength{\tabcolsep}{2pt}
\begin{longtable}{@{}l l rrr@{\hspace{1em}}rrr@{\hspace{1em}}rrr@{\hspace{1em}}rrr@{}}
\caption{Full experimental results for all models, configurations, and thinking modes.\\
\colorbox{gray!25}{Shaded} rows show freeform baseline accuracy (\%); unshaded rows show change from baseline (negative = degradation).\\
Accuracy is lenient (answer correct; format validity not required). Config notation: G = grammar in prompt, E = few-shot examples, T = task description, C = constrained decoding.}\\
\label{tab:format-tax-consolidated}\\
\toprule
 &  & \multicolumn{3}{c}{1-Turn} & \multicolumn{3}{c}{1-Turn + Think} & \multicolumn{3}{c}{2-Turn} & \multicolumn{3}{c}{2-Turn + Think} \\
\cmidrule(lr){3-5} \cmidrule(lr){6-8} \cmidrule(lr){9-11} \cmidrule(lr){12-14}
Config & Format & GPQA & Math & Zebra & GPQA & Math & Zebra & GPQA & Math & Zebra & GPQA & Math & Zebra \\
\midrule
\endfirsthead
\multicolumn{14}{c}{\small\textit{Table \ref{tab:format-tax-consolidated} continued from previous page}} \\[0.5em]
\toprule
 &  & \multicolumn{3}{c}{1-Turn} & \multicolumn{3}{c}{1-Turn + Think} & \multicolumn{3}{c}{2-Turn} & \multicolumn{3}{c}{2-Turn + Think} \\
\cmidrule(lr){3-5} \cmidrule(lr){6-8} \cmidrule(lr){9-11} \cmidrule(lr){12-14}
Config & Format & GPQA & Math & Zebra & GPQA & Math & Zebra & GPQA & Math & Zebra & GPQA & Math & Zebra \\
\midrule
\endhead
\multicolumn{14}{r}{\small\textit{Continued on next page}} \\
\endfoot
\bottomrule
\endlastfoot
\rowcolor{gray!25} \multicolumn{2}{l}{\qwenicon~\large\textbf{qwen3-32b}} & 52.0 & 84.4 & 64.8 & 52.2 & 93.7 & 77.0 & 52.0 & 84.4 & 64.8 & 52.2 & 93.7 & 77.0 \\
\multirow{4}{*}{\texttt{G·T·}} & JSON & \ccol{FCF3F3} -4.0 & \ccol{FBEDED} -6.4 & \ccol{FCF5F5} -3.6 & \ccol{D4ECD4} 6.9 & \ccol{F7FBF7} 1.3 & \ccol{F1F8F1} 2.2 & \ccol{F8FCF8} 1.0 & \ccol{F1F8F1} 2.2 & \ccol{F8FCF8} 1.0 & \ccol{CEE9CE} 7.9 & \ccol{F3F9F3} 1.9 & \ccol{C0E2C0} 10.2 \\*
 & XML & \ccol{FBEFEF} -5.6 & \ccol{FBEDED} -6.4 & \ccol{FBECEC} -6.8 & 0.3 & 0.7 & \ccol{F6FBF6} 1.4 & 0.5 & \ccol{F4FAF4} 1.8 & \ccol{F3FAF3} 1.8 & \ccol{C1E3C1} 9.9 & \ccol{F5FAF5} 1.5 & \ccol{CFE9CF} 7.8 \\*
 & Markdown & \ccol{FEFAFA} -1.5 & \ccol{FCF4F4} -3.7 & \ccol{FAE9E9} -8.0 & \ccol{FEFAFA} -1.7 & 0.3 & \ccol{F8FCF8} 1.0 & \ccol{F8FCF8} 1.0 & \ccol{F0F8F0} 2.3 & \ccol{F8FCF8} 1.0 & \ccol{B8DFB8} 11.4 & \ccol{EEF7EE} 2.7 & \ccol{C3E4C3} 9.6 \\*
 & LaTeX & \ccol{FCF5F5} -3.5 & \ccol{FCF2F2} -4.7 & \ccol{FCF2F2} -4.4 & \ccol{FBEDED} -6.2 & -0.1 & -0.4 & \ccol{F8FCF8} 1.0 & \ccol{F3F9F3} 1.9 & 0.2 & \ccol{D1EAD1} 7.4 & \ccol{F3F9F3} 1.9 & \ccol{C0E2C0} 10.2 \\
\noalign{\vspace{0.7em}}\nopagebreak
\multirow{4}{*}{\texttt{·ET·}} & JSON & \ccol{FAE7E7} -8.6 & \ccol{F9E6E6} -8.8 & \ccol{F9E4E4} -9.8 & \ccol{EDF7ED} 2.9 & \ccol{F8FCF8} 1.1 & \ccol{F6FBF6} 1.4 & \ccol{F8FCF8} 1.0 & \ccol{F2F9F2} 2.0 & 0.4 & \ccol{BEE2BE} 10.4 & \ccol{F3F9F3} 1.9 & \ccol{BBE0BB} 11.0 \\*
 & XML & \ccol{FAE7E7} -8.6 & \ccol{FBEEEE} -5.9 & \ccol{FAEBEB} -7.0 & \ccol{FEFAFA} -1.7 & 0.3 & \ccol{F5FAF5} 1.6 & 0.5 & \ccol{F2F9F2} 2.0 & 0.2 & \ccol{DDF0DD} 5.4 & \ccol{EFF8EF} 2.5 & \ccol{C2E4C2} 9.8 \\*
 & Markdown & \ccol{FCF1F1} -5.0 & \ccol{FBF0F0} -5.4 & \ccol{FBECEC} -6.6 & \ccol{FDF7F7} -2.7 & 0.3 & -0.4 & \ccol{F8FCF8} 1.0 & \ccol{EEF7EE} 2.6 & 0.2 & \ccol{D1EAD1} 7.4 & \ccol{EEF7EE} 2.7 & \ccol{B7DFB7} 11.6 \\*
 & LaTeX & \ccol{FAEBEB} -7.1 & \ccol{FCF4F4} -3.8 & \ccol{FAEAEA} -7.4 & \ccol{FBEDED} -6.2 & 0.9 & \ccol{F3FAF3} 1.8 & 0.5 & \ccol{F1F8F1} 2.2 & 0.2 & \ccol{D4ECD4} 6.9 & \ccol{F1F8F1} 2.2 & \ccol{C0E2C0} 10.2 \\
\noalign{\vspace{0.7em}}\nopagebreak
\multirow{4}{*}{\texttt{GET·}} & JSON & \ccol{FAE7E7} -8.6 & \ccol{FBF0F0} -5.3 & \ccol{FBEDED} -6.4 & \ccol{FDF9F9} -2.2 & -0.3 & \ccol{F7FBF7} 1.2 & \ccol{F8FCF8} 1.0 & \ccol{F1F9F1} 2.2 & 0.4 & \ccol{D4ECD4} 6.9 & \ccol{F3F9F3} 1.9 & \ccol{BDE1BD} 10.6 \\*
 & XML & \ccol{FAE8E8} -8.1 & \ccol{FAEBEB} -7.3 & \ccol{FAE8E8} -8.2 & -0.7 & -0.1 & -0.4 & \ccol{F5FAF5} 1.5 & \ccol{F0F8F0} 2.4 & 0.8 & \ccol{D4ECD4} 6.9 & \ccol{F0F8F0} 2.3 & \ccol{BBE0BB} 11.0 \\*
 & Markdown & \ccol{FCF2F2} -4.5 & \ccol{FCF5F5} -3.5 & \ccol{FDF7F7} -2.8 & \ccol{FDF7F7} -2.7 & 0.7 & \ccol{ECF6EC} 3.0 & \ccol{F5FAF5} 1.5 & \ccol{EEF7EE} 2.7 & 0.4 & \ccol{C1E3C1} 9.9 & \ccol{F2F9F2} 2.1 & \ccol{C5E5C5} 9.4 \\*
 & LaTeX & \ccol{FCF2F2} -4.5 & \ccol{FCF5F5} -3.6 & \ccol{FCF4F4} -4.0 & \ccol{FBEDED} -6.2 & -0.7 & \ccol{F7FBF7} 1.2 & \ccol{F8FCF8} 1.0 & \ccol{EDF6ED} 2.9 & 0.8 & \ccol{DDF0DD} 5.4 & \ccol{F2F9F2} 2.1 & \ccol{BFE2BF} 10.4 \\
\noalign{\vspace{0.7em}}\nopagebreak
\multirow{4}{*}{\texttt{G·TC}} & JSON & \ccol{FAE8E8} -8.1 & \ccol{F8E2E2} -10.6 & \ccol{FBEEEE} -6.2 & 0.3 & -0.7 & \ccol{F2F9F2} 2.0 & \ccol{F8FCF8} 1.0 & \ccol{F6FBF6} 1.4 & 0.8 & \ccol{BEE2BE} 10.4 & \ccol{F0F8F0} 2.3 & \ccol{BFE2BF} 10.4 \\*
 & XML & \ccol{FBEEEE} -6.1 & \ccol{F9E5E5} -9.2 & \ccol{FBF0F0} -5.2 & \ccol{F6FBF6} 1.3 & -0.3 & 0.6 & \ccol{EFF8EF} 2.5 & 0.8 & 0.4 & \ccol{BEE2BE} 10.4 & \ccol{FCF1F1} -5.1 & \ccol{C7E6C7} 9.0 \\*
 & Markdown & \ccol{FBEFEF} -5.6 & \ccol{FAE8E8} -8.2 & \ccol{FAE7E7} -8.4 & \ccol{F6FBF6} 1.3 & \ccol{FDF8F8} -2.4 & 0.4 & \ccol{EFF8EF} 2.5 & 0.6 & 0.6 & \ccol{C8E6C8} 8.9 & \ccol{F8FCF8} 1.0 & \ccol{CCE8CC} 8.2 \\*
 & LaTeX & \ccol{FCF5F5} -3.5 & -0.9 & \ccol{FCF2F2} -4.4 & \ccol{FDF9F9} -2.2 & \ccol{FDF9F9} -2.1 & -0.6 & \ccol{F8FCF8} 1.0 & \ccol{F5FAF5} 1.5 & \ccol{F3FAF3} 1.8 & \ccol{C5E5C5} 9.4 & \ccol{F2F9F2} 2.1 & \ccol{D0EAD0} 7.6 \\
\noalign{\vspace{0.7em}}\nopagebreak
\multirow{4}{*}{\texttt{·ETC}} & JSON & \ccol{FAE8E8} -8.1 & \ccol{F4CECE} -17.8 & \ccol{FAEAEA} -7.4 & 0.8 & -0.7 & \ccol{F7FBF7} 1.2 & \ccol{F8FCF8} 1.0 & 0.6 & 0.4 & \ccol{C5E5C5} 9.4 & \ccol{F7FBF7} 1.3 & \ccol{BDE1BD} 10.6 \\*
 & XML & \ccol{F8E0E0} -11.1 & \ccol{F8DEDE} -12.0 & \ccol{FBEFEF} -5.6 & -0.7 & 0.1 & -0.6 & \ccol{ECF6EC} 3.0 & \ccol{F5FAF5} 1.6 & 0.2 & \ccol{C8E6C8} 8.9 & \ccol{F4FAF4} 1.7 & \ccol{C0E2C0} 10.2 \\*
 & Markdown & \ccol{FBEEEE} -6.1 & \ccol{F7DCDC} -12.6 & \ccol{FAE7E7} -8.4 & 0.8 & 0.3 & \ccol{F5FAF5} 1.6 & \ccol{F8FCF8} 1.0 & \ccol{F5FAF5} 1.6 & \ccol{F0F8F0} 2.4 & \ccol{D7EDD7} 6.4 & \ccol{F6FAF6} 1.5 & \ccol{C5E5C5} 9.4 \\*
 & LaTeX & \ccol{FBEDED} -6.6 & \ccol{FCF2F2} -4.4 & \ccol{F9E3E3} -10.0 & \ccol{FEFAFA} -1.7 & \ccol{FEFBFB} -1.3 & \ccol{F6FBF6} 1.4 & \ccol{F8FCF8} 1.0 & \ccol{F4FAF4} 1.8 & 0.4 & \ccol{D1EAD1} 7.4 & \ccol{F7FBF7} 1.3 & \ccol{C2E4C2} 9.8 \\
\noalign{\vspace{0.7em}}\nopagebreak
\multirow{4}{*}{\texttt{GETC}} & JSON & \ccol{FBEEEE} -6.1 & \ccol{F9E4E4} -9.8 & \ccol{FAEAEA} -7.6 & -0.7 & -0.1 & -0.4 & \ccol{ECF6EC} 3.0 & 1.0 & \ccol{F8FCF8} 1.0 & \ccol{C1E3C1} 9.9 & 0.7 & \ccol{B8DFB8} 11.4 \\*
 & XML & \ccol{FBEDED} -6.6 & \ccol{F8E0E0} -11.3 & \ccol{F8E0E0} -11.2 & -0.7 & -0.3 & 0.2 & \ccol{F5FAF5} 1.5 & \ccol{F6FBF6} 1.4 & 0.2 & \ccol{BEE2BE} 10.4 & \ccol{F7FBF7} 1.3 & \ccol{C0E2C0} 10.2 \\*
 & Markdown & \ccol{F9E6E6} -9.1 & \ccol{F9E5E5} -9.4 & \ccol{F7D9D9} -13.6 & \ccol{FCF2F2} -4.7 & 0.2 & \ccol{F2F9F2} 2.0 & \ccol{E9F5E9} 3.5 & 0.6 & 0.6 & \ccol{CBE7CB} 8.4 & \ccol{EFF8EF} 2.5 & \ccol{C6E5C6} 9.2 \\*
 & LaTeX & \ccol{FAEAEA} -7.6 & \ccol{FDF6F6} -3.0 & \ccol{FBEDED} -6.4 & \ccol{FBEFEF} -5.7 & -0.4 & \ccol{FEFCFC} -1.0 & 0.0 & \ccol{EEF7EE} 2.7 & \ccol{F8FCF8} 1.0 & \ccol{DDF0DD} 5.4 & \ccol{F7FBF7} 1.3 & \ccol{BAE0BA} 11.2 \\
\noalign{\vspace{1em}}
\rowcolor{gray!25} \multicolumn{2}{l}{\qwenicon~\large\textbf{qwen3-8b}} & 40.6 & 84.0 & 63.1 & 43.1 & 91.9 & 73.3 & 40.6 & 84.0 & 63.1 & 43.1 & 91.9 & 73.3 \\
\multirow{4}{*}{\texttt{G·T·}} & JSON & \ccol{FBEFEF} -5.7 & \ccol{F9E5E5} -9.4 & \ccol{FAE8E8} -8.3 & \ccol{D4ECD4} 6.9 & \ccol{F2F9F2} 2.1 & \ccol{F2F9F2} 2.1 & \ccol{C8E6C8} 8.9 & 0.4 & \ccol{E8F4E8} 3.7 & \ccol{AFDBAF} 13.0 & \ccol{EAF6EA} 3.3 & \ccol{C2E3C2} 9.9 \\*
 & XML & \ccol{F0F8F0} 2.4 & \ccol{FCF3F3} -4.2 & \ccol{F9E6E6} -8.9 & \ccol{E4F2E4} 4.4 & \ccol{EFF8EF} 2.5 & \ccol{F3F9F3} 1.9 & \ccol{C8E6C8} 8.9 & 0.0 & \ccol{E9F5E9} 3.5 & \ccol{BBE0BB} 10.9 & \ccol{E9F5E9} 3.5 & \ccol{BFE2BF} 10.3 \\*
 & Markdown & \ccol{FEFBFB} -1.2 & \ccol{FCF4F4} -3.7 & \ccol{FBF0F0} -5.5 & \ccol{E7F4E7} 3.9 & \ccol{F1F8F1} 2.3 & \ccol{F2F9F2} 2.1 & \ccol{D7EDD7} 6.4 & 0.5 & \ccol{E8F4E8} 3.7 & \ccol{AFDBAF} 13.0 & \ccol{E5F3E5} 4.1 & \ccol{C1E3C1} 10.1 \\*
 & LaTeX & \ccol{F6FBF6} 1.4 & \ccol{FBEFEF} -5.6 & \ccol{ECA9A9} -31.3 & \ccol{FEFAFA} -1.7 & \ccol{FEFBFB} -1.2 & \ccol{F8FCF8} 1.1 & \ccol{D1EAD1} 7.4 & \ccol{F8E1E1} -10.6 & \ccol{EA9B9B} -36.3 & \ccol{C4E5C4} 9.4 & -0.3 & \ccol{F9E5E5} -9.3 \\
\noalign{\vspace{0.7em}}\nopagebreak
\multirow{4}{*}{\texttt{·ET·}} & JSON & \ccol{F9E7E7} -8.8 & \ccol{FBEDED} -6.4 & \ccol{FAEBEB} -7.1 & \ccol{F6FBF6} 1.3 & \ccol{EEF7EE} 2.7 & \ccol{F8FCF8} 1.1 & \ccol{D7EDD7} 6.4 & 0.6 & \ccol{E1F1E1} 4.7 & \ccol{BBE0BB} 10.9 & \ccol{EDF7ED} 2.9 & \ccol{B2DCB2} 12.5 \\*
 & XML & \ccol{FBEFEF} -5.7 & \ccol{FCF5F5} -3.6 & \ccol{FBEEEE} -5.9 & \ccol{EDF7ED} 2.9 & \ccol{F2F9F2} 2.1 & \ccol{F2F9F2} 2.1 & \ccol{C8E6C8} 8.9 & -0.2 & \ccol{E0F1E0} 4.9 & \ccol{C1E3C1} 9.9 & \ccol{EEF7EE} 2.6 & \ccol{B7DEB7} 11.7 \\*
 & Markdown & \ccol{FEFAFA} -1.7 & \ccol{FDF6F6} -3.3 & \ccol{FAEBEB} -7.3 & 0.3 & 0.9 & 0.1 & \ccol{D4ECD4} 6.9 & 0.2 & \ccol{E8F4E8} 3.7 & \ccol{CEE9CE} 7.9 & \ccol{E3F2E3} 4.5 & \ccol{C7E6C7} 9.1 \\*
 & LaTeX & \ccol{FCF3F3} -4.2 & \ccol{FDF6F6} -3.1 & \ccol{F9E5E5} -9.5 & \ccol{F0F8F0} 2.4 & \ccol{F2F9F2} 2.1 & 0.5 & \ccol{D7EDD7} 6.4 & 0.2 & \ccol{E3F2E3} 4.5 & \ccol{BBE0BB} 10.9 & \ccol{E5F3E5} 4.1 & \ccol{B4DDB4} 12.1 \\
\noalign{\vspace{0.7em}}\nopagebreak
\multirow{4}{*}{\texttt{GET·}} & JSON & \ccol{FAE8E8} -8.3 & \ccol{F8DFDF} -11.5 & \ccol{F9E4E4} -9.7 & \ccol{FEFBFB} -1.2 & \ccol{F8FCF8} 1.1 & \ccol{F2F9F2} 2.1 & \ccol{D7EDD7} 6.4 & -0.2 & \ccol{DFF0DF} 5.1 & \ccol{B2DCB2} 12.5 & \ccol{E7F4E7} 3.9 & \ccol{C6E5C6} 9.3 \\*
 & XML & \ccol{FCF2F2} -4.7 & \ccol{FBEEEE} -6.0 & \ccol{F8E0E0} -11.1 & \ccol{EDF7ED} 2.9 & \ccol{F7FBF7} 1.3 & \ccol{F3F9F3} 1.9 & \ccol{C8E6C8} 8.9 & 0.2 & \ccol{E1F1E1} 4.7 & \ccol{BEE2BE} 10.4 & \ccol{E8F4E8} 3.7 & \ccol{BEE2BE} 10.5 \\*
 & Markdown & \ccol{F9E5E5} -9.3 & \ccol{FBEEEE} -5.9 & \ccol{F9E5E5} -9.5 & \ccol{F3F9F3} 1.9 & \ccol{E9F5E9} 3.5 & \ccol{FDF9F9} -2.1 & \ccol{D7EDD7} 6.4 & 0.4 & \ccol{E9F5E9} 3.5 & \ccol{BEE2BE} 10.4 & \ccol{E7F4E7} 3.8 & \ccol{BEE2BE} 10.5 \\*
 & LaTeX & \ccol{FEFAFA} -1.7 & \ccol{FBEFEF} -5.8 & \ccol{F9E4E4} -9.7 & \ccol{DDF0DD} 5.4 & \ccol{F8FCF8} 1.0 & \ccol{F7FBF7} 1.3 & \ccol{DAEEDA} 5.9 & 0.0 & \ccol{EAF5EA} 3.3 & \ccol{C1E3C1} 9.9 & \ccol{E9F5E9} 3.5 & \ccol{AEDBAE} 13.1 \\
\noalign{\vspace{0.7em}}\nopagebreak
\multirow{4}{*}{\texttt{G·TC}} & JSON & \ccol{FBECEC} -6.7 & \ccol{F5D2D2} -16.3 & \ccol{FAEBEB} -7.3 & \ccol{F6FBF6} 1.3 & \ccol{F5FAF5} 1.5 & \ccol{E7F4E7} 3.9 & \ccol{E0F1E0} 4.9 & \ccol{FDF7F7} -2.6 & \ccol{E1F1E1} 4.7 & \ccol{B8DFB8} 11.5 & \ccol{F3F9F3} 1.9 & \ccol{BEE2BE} 10.5 \\*
 & XML & \ccol{FAE9E9} -7.7 & \ccol{F9E3E3} -10.2 & \ccol{FCF2F2} -4.7 & \ccol{DDF0DD} 5.4 & -0.9 & \ccol{F4FAF4} 1.7 & \ccol{D7EDD7} 6.4 & \ccol{FCF5F5} -3.6 & \ccol{E3F2E3} 4.5 & \ccol{BBE0BB} 10.9 & \ccol{FDF7F7} -2.7 & \ccol{C2E3C2} 9.9 \\*
 & Markdown & \ccol{F6FBF6} 1.4 & \ccol{FBEEEE} -6.0 & \ccol{FBEDED} -6.5 & 0.8 & 0.7 & \ccol{F3F9F3} 1.9 & \ccol{D7EDD7} 6.4 & -1.0 & \ccol{E6F4E6} 3.9 & \ccol{C4E5C4} 9.4 & \ccol{E7F4E7} 3.9 & \ccol{D8EDD8} 6.3 \\*
 & LaTeX & \ccol{FDF7F7} -2.7 & \ccol{FCF2F2} -4.6 & \ccol{EA9E9E} -35.3 & \ccol{FEFBFB} -1.2 & \ccol{FDF7F7} -2.6 & \ccol{FEFBFB} -1.3 & \ccol{D7EDD7} 6.4 & \ccol{F9E4E4} -9.6 & \ccol{EFB5B5} -26.9 & \ccol{DDF0DD} 5.4 & \ccol{FCF3F3} -4.4 & \ccol{F5D0D0} -16.9 \\
\noalign{\vspace{0.7em}}\nopagebreak
\multirow{4}{*}{\texttt{·ETC}} & JSON & \ccol{F8DEDE} -11.8 & \ccol{F6D8D8} -13.9 & \ccol{F8DDDD} -12.1 & \ccol{FEFAFA} -1.7 & \ccol{FDFAFA} -1.7 & \ccol{F4FAF4} 1.7 & \ccol{DDF0DD} 5.4 & -1.0 & \ccol{E6F4E6} 3.9 & \ccol{D1EAD1} 7.4 & 0.4 & \ccol{B9E0B9} 11.3 \\*
 & XML & \ccol{FBEDED} -6.2 & \ccol{F9E6E6} -9.0 & \ccol{FAE8E8} -8.1 & \ccol{EDF7ED} 2.9 & \ccol{F2F9F2} 2.1 & \ccol{F8FCF8} 1.1 & \ccol{D7EDD7} 6.4 & 0.4 & \ccol{E1F1E1} 4.7 & \ccol{C4E5C4} 9.4 & \ccol{F4FAF4} 1.7 & \ccol{B9E0B9} 11.3 \\*
 & Markdown & \ccol{FDF7F7} -2.7 & \ccol{FAE9E9} -7.7 & \ccol{FAE7E7} -8.5 & \ccol{F3F9F3} 1.9 & 0.1 & \ccol{F8FCF8} 1.1 & \ccol{D7EDD7} 6.4 & 0.6 & \ccol{E0F1E0} 4.9 & \ccol{CBE7CB} 8.4 & \ccol{E9F5E9} 3.5 & \ccol{BEE2BE} 10.5 \\*
 & LaTeX & \ccol{FEFAFA} -1.7 & \ccol{FBECEC} -6.8 & \ccol{FAEAEA} -7.7 & \ccol{E4F2E4} 4.4 & 0.5 & \ccol{F7FBF7} 1.3 & \ccol{D7EDD7} 6.4 & 0.2 & \ccol{E0F1E0} 4.9 & \ccol{CEE9CE} 7.9 & \ccol{ECF6EC} 3.1 & \ccol{B8DFB8} 11.5 \\
\noalign{\vspace{0.7em}}\nopagebreak
\multirow{4}{*}{\texttt{GETC}} & JSON & \ccol{F7DDDD} -12.3 & \ccol{F5D1D1} -16.7 & \ccol{F8DEDE} -11.9 & \ccol{FEFAFA} -1.7 & \ccol{F2F9F2} 2.1 & \ccol{F3F9F3} 1.9 & \ccol{E0F1E0} 4.9 & \ccol{FDF9F9} -2.0 & \ccol{E0F1E0} 4.9 & \ccol{C4E5C4} 9.4 & -0.7 & \ccol{BCE1BC} 10.9 \\*
 & XML & \ccol{FBF0F0} -5.2 & \ccol{F9E6E6} -8.9 & \ccol{F8E0E0} -11.1 & \ccol{EAF5EA} 3.4 & \ccol{EDF7ED} 2.9 & \ccol{F1F8F1} 2.3 & \ccol{D7EDD7} 6.4 & 0.0 & \ccol{E1F1E1} 4.7 & \ccol{CBE7CB} 8.4 & \ccol{EAF6EA} 3.3 & \ccol{C4E4C4} 9.5 \\*
 & Markdown & \ccol{FAE9E9} -7.7 & \ccol{F8E1E1} -10.9 & \ccol{F8DFDF} -11.5 & \ccol{FEFBFB} -1.2 & \ccol{F7FBF7} 1.3 & \ccol{F1F8F1} 2.3 & \ccol{D7EDD7} 6.4 & 0.2 & \ccol{E1F1E1} 4.7 & \ccol{CEE9CE} 7.9 & \ccol{ECF6EC} 3.1 & \ccol{C6E5C6} 9.3 \\*
 & LaTeX & -0.2 & \ccol{FCF2F2} -4.6 & \ccol{F9E4E4} -9.7 & \ccol{E4F2E4} 4.4 & 0.1 & \ccol{F4FAF4} 1.7 & \ccol{DDF0DD} 5.4 & 0.2 & \ccol{E1F1E1} 4.7 & \ccol{B5DEB5} 12.0 & \ccol{E7F4E7} 3.8 & \ccol{B3DDB3} 12.3 \\
\noalign{\vspace{1em}}
\rowcolor{gray!25} \multicolumn{2}{l}{\allenaiicon~\large\textbf{olmo3.1-32b}} & 53.7 & 91.4 & 72.9 & 42.9 & 91.9 & 69.3 & 53.7 & 91.4 & 72.9 & 42.9 & 91.9 & 69.3 \\
\multirow{4}{*}{\texttt{G·T·}} & JSON & -0.7 & \ccol{FCF1F1} -4.8 & \ccol{E2F2E2} 4.7 & \ccol{ECF6EC} 3.0 & 0.9 & 0.9 & \ccol{CEE9CE} 7.9 & \ccol{EAF5EA} 3.4 & 0.5 & \ccol{AEDAAE} 13.1 & \ccol{F9E4E4} -9.7 & \ccol{AFDBAF} 12.9 \\*
 & XML & -0.7 & \ccol{FCF2F2} -4.6 & \ccol{DDEFDD} 5.5 & \ccol{F5FAF5} 1.5 & \ccol{F6FBF6} 1.3 & -0.3 & \ccol{9CD39C} 16.0 & \ccol{ECF6EC} 3.0 & \ccol{FEFAFA} -1.5 & \ccol{CDE8CD} 8.1 & \ccol{F5D4D4} -15.7 & \ccol{CAE7CA} 8.5 \\*
 & Markdown & -0.7 & \ccol{FBECEC} -6.8 & \ccol{EDF7ED} 2.9 & \ccol{E3F2E3} 4.5 & -0.7 & 0.5 & \ccol{CEE9CE} 7.9 & \ccol{EAF5EA} 3.4 & 0.5 & \ccol{A8D8A8} 14.1 & \ccol{F4CBCB} -18.9 & \ccol{B8DFB8} 11.5 \\*
 & LaTeX & -0.7 & \ccol{FBEEEE} -6.2 & \ccol{D9EED9} 6.1 & \ccol{CDE8CD} 8.1 & \ccol{F9E3E3} -9.9 & \ccol{FDF6F6} -3.3 & \ccol{DDF0DD} 5.4 & \ccol{EBF6EB} 3.2 & \ccol{E1F1E1} 4.9 & \ccol{C0E3C0} 10.1 & \ccol{F2C4C4} -21.5 & \ccol{CFE9CF} 7.7 \\
\noalign{\vspace{0.7em}}\nopagebreak
\multirow{4}{*}{\texttt{·ET·}} & JSON & \ccol{FCF3F3} -4.2 & \ccol{FDF7F7} -2.6 & -0.5 & 0.0 & -0.1 & \ccol{FEFAFA} -1.5 & \ccol{D1EAD1} 7.4 & \ccol{ECF6EC} 3.0 & -0.3 & \ccol{DFF1DF} 5.1 & \ccol{E9F5E9} 3.5 & \ccol{B3DDB3} 12.3 \\*
 & XML & -0.7 & 0.6 & \ccol{D8EDD8} 6.3 & \ccol{FDF6F6} -3.0 & 0.7 & -0.9 & \ccol{CEE9CE} 7.9 & \ccol{EAF5EA} 3.4 & \ccol{F8FCF8} 1.1 & \ccol{C7E5C7} 9.1 & \ccol{F1F9F1} 2.1 & \ccol{B6DEB6} 11.7 \\*
 & Markdown & 0.8 & \ccol{FDF7F7} -2.6 & \ccol{FCF5F5} -3.5 & \ccol{FCF3F3} -4.0 & \ccol{FDF9F9} -2.1 & \ccol{FDF8F8} -2.5 & \ccol{B8DFB8} 11.4 & \ccol{E8F5E8} 3.6 & \ccol{F2F9F2} 2.1 & \ccol{C7E5C7} 9.1 & \ccol{F4FAF4} 1.7 & \ccol{ADDAAD} 13.3 \\*
 & LaTeX & \ccol{FBEDED} -6.2 & 0.4 & \ccol{ECF6EC} 3.1 & \ccol{FDF9F9} -2.0 & -0.9 & \ccol{FDF9F9} -2.1 & \ccol{B8DFB8} 11.4 & \ccol{E7F4E7} 3.8 & \ccol{F3F9F3} 1.9 & \ccol{C7E5C7} 9.1 & -0.5 & \ccol{B8DFB8} 11.5 \\
\noalign{\vspace{0.7em}}\nopagebreak
\multirow{4}{*}{\texttt{GET·}} & JSON & \ccol{F9E5E5} -9.3 & \ccol{FBECEC} -6.6 & \ccol{DAEEDA} 5.9 & \ccol{E6F3E6} 4.0 & -0.1 & \ccol{FEFCFC} -1.1 & \ccol{D7EDD7} 6.4 & \ccol{E7F4E7} 3.8 & -0.1 & \ccol{C0E3C0} 10.1 & 0.3 & \ccol{B8DFB8} 11.5 \\*
 & XML & \ccol{FDF6F6} -3.2 & \ccol{FAE8E8} -8.2 & \ccol{E2F2E2} 4.7 & \ccol{DCEFDC} 5.6 & 0.7 & -0.5 & \ccol{EAF5EA} 3.4 & \ccol{E7F4E7} 3.8 & 0.7 & \ccol{CAE7CA} 8.6 & \ccol{F6D9D9} -13.9 & \ccol{D3EBD3} 7.1 \\*
 & Markdown & -0.7 & \ccol{FDF5F5} -3.4 & \ccol{EAF6EA} 3.3 & \ccol{FDF6F6} -3.0 & \ccol{FCF4F4} -3.9 & -0.5 & \ccol{DDF0DD} 5.4 & \ccol{E7F4E7} 3.8 & \ccol{E9F5E9} 3.5 & \ccol{E3F2E3} 4.5 & \ccol{F7DBDB} -12.9 & \ccol{E4F3E4} 4.3 \\*
 & LaTeX & -0.7 & \ccol{FBECEC} -6.8 & \ccol{D3EBD3} 7.1 & \ccol{FEFAFA} -1.7 & \ccol{FAEBEB} -7.1 & \ccol{FEFCFC} -1.1 & \ccol{F0F8F0} 2.4 & \ccol{EBF6EB} 3.2 & \ccol{F2F9F2} 2.1 & \ccol{D3EBD3} 7.1 & \ccol{F4CDCD} -17.9 & \ccol{C4E4C4} 9.5 \\
\noalign{\vspace{0.7em}}\nopagebreak
\multirow{4}{*}{\texttt{G·TC}} & JSON & \ccol{FEFBFB} -1.2 & \ccol{FCF3F3} -4.0 & \ccol{F3F9F3} 1.9 & \ccol{E9F5E9} 3.5 & 0.9 & 0.1 & \ccol{D4ECD4} 6.9 & \ccol{E8F5E8} 3.6 & 0.3 & \ccol{B7DFB7} 11.6 & \ccol{FBF0F0} -5.3 & \ccol{AFDBAF} 12.9 \\*
 & XML & 0.3 & \ccol{F6D8D8} -14.1 & \ccol{FDF9F9} -2.1 & -0.5 & 0.1 & 0.3 & \ccol{BEE2BE} 10.4 & \ccol{FEFAFA} -1.6 & -0.7 & \ccol{CDE8CD} 8.1 & \ccol{F4CDCD} -18.1 & \ccol{D9EED9} 6.1 \\*
 & Markdown & 0.3 & \ccol{F9E6E6} -8.8 & 0.3 & \ccol{D9EED9} 6.1 & 0.3 & -0.1 & \ccol{CEE9CE} 7.9 & \ccol{F1F8F1} 2.2 & -0.1 & \ccol{A8D8A8} 14.1 & \ccol{F3CACA} -19.1 & \ccol{C5E5C5} 9.3 \\*
 & LaTeX & \ccol{FEFBFB} -1.2 & \ccol{FBEFEF} -5.8 & \ccol{EAF6EA} 3.3 & \ccol{CAE7CA} 8.6 & \ccol{F8E2E2} -10.5 & 0.1 & \ccol{AFDBAF} 13.0 & \ccol{E6F4E6} 4.0 & \ccol{F1F8F1} 2.3 & \ccol{98D198} 16.7 & \ccol{F4CBCB} -18.9 & \ccol{BCE1BC} 10.7 \\
\noalign{\vspace{0.7em}}\nopagebreak
\multirow{4}{*}{\texttt{·ETC}} & JSON & \ccol{F8DEDE} -11.8 & \ccol{FDF8F8} -2.2 & 0.1 & \ccol{FDF8F8} -2.5 & 0.1 & -0.7 & \ccol{D7EDD7} 6.4 & \ccol{F0F8F0} 2.4 & \ccol{F2F9F2} 2.1 & \ccol{C7E5C7} 9.1 & \ccol{F4FAF4} 1.7 & \ccol{A9D8A9} 13.9 \\*
 & XML & \ccol{FCF2F2} -4.7 & \ccol{FAE8E8} -8.2 & \ccol{FEFBFB} -1.1 & -0.5 & \ccol{FEFCFC} -1.1 & 0.5 & \ccol{D7EDD7} 6.4 & 0.4 & 0.7 & \ccol{D9EED9} 6.1 & \ccol{EEF7EE} 2.7 & \ccol{B6DEB6} 11.7 \\*
 & Markdown & \ccol{F0F8F0} 2.4 & \ccol{FDF8F8} -2.4 & \ccol{EDF7ED} 2.9 & \ccol{FEFAFA} -1.5 & -0.1 & \ccol{FDF8F8} -2.3 & \ccol{F0F8F0} 2.4 & \ccol{EAF5EA} 3.4 & \ccol{EAF6EA} 3.3 & \ccol{CAE7CA} 8.6 & \ccol{F4FAF4} 1.7 & \ccol{ABD9AB} 13.5 \\*
 & LaTeX & \ccol{F3F9F3} 1.9 & \ccol{FEFBFB} -1.4 & \ccol{CCE8CC} 8.3 & \ccol{EFF8EF} 2.5 & -0.3 & -0.1 & \ccol{D4ECD4} 6.9 & \ccol{E8F5E8} 3.6 & 0.5 & \ccol{C0E3C0} 10.1 & -0.9 & \ccol{BCE1BC} 10.7 \\
\noalign{\vspace{0.7em}}\nopagebreak
\multirow{4}{*}{\texttt{GETC}} & JSON & \ccol{FCF3F3} -4.2 & \ccol{FBEFEF} -5.6 & \ccol{E5F3E5} 4.1 & \ccol{E9F5E9} 3.5 & \ccol{F6FBF6} 1.3 & \ccol{FDF8F8} -2.3 & \ccol{B8DFB8} 11.4 & \ccol{EBF6EB} 3.2 & \ccol{E3F2E3} 4.5 & \ccol{BAE0BA} 11.1 & -0.7 & \ccol{AFDBAF} 12.9 \\*
 & XML & \ccol{FAE9E9} -7.7 & \ccol{F7D9D9} -13.6 & -0.3 & \ccol{E6F3E6} 4.0 & 0.9 & -0.9 & \ccol{D7EDD7} 6.4 & 0.4 & \ccol{EDF7ED} 2.9 & \ccol{D3EBD3} 7.1 & \ccol{F5D4D4} -15.7 & \ccol{C1E3C1} 9.9 \\*
 & Markdown & -0.7 & \ccol{FAEBEB} -7.2 & \ccol{F7FBF7} 1.3 & \ccol{FDF9F9} -2.0 & \ccol{FBF0F0} -5.3 & \ccol{FEFCFC} -1.1 & \ccol{D1EAD1} 7.4 & \ccol{E5F3E5} 4.2 & \ccol{FEFBFB} -1.1 & \ccol{D9EED9} 6.1 & \ccol{F3C8C8} -19.9 & \ccol{CFE9CF} 7.7 \\*
 & LaTeX & \ccol{FDF7F7} -2.7 & \ccol{FBECEC} -6.7 & \ccol{C6E5C6} 9.3 & \ccol{EFF8EF} 2.5 & \ccol{FAEBEB} -7.3 & \ccol{FEFAFA} -1.5 & \ccol{DAEEDA} 5.9 & \ccol{EAF5EA} 3.4 & \ccol{F5FAF5} 1.5 & \ccol{AEDAAE} 13.1 & \ccol{F3CACA} -19.3 & \ccol{D0EAD0} 7.5 \\
\noalign{\vspace{1em}}
\rowcolor{gray!25} \multicolumn{2}{l}{\allenaiicon~\large\textbf{olmo3-7b}} & 43.3 & 90.5 & 69.3 & 31.8 & 89.7 & 51.7 & 43.3 & 90.5 & 69.3 & 31.8 & 89.7 & 51.7 \\
\multirow{4}{*}{\texttt{G·T·}} & JSON & \ccol{F9E5E5} -9.4 & \ccol{FBEEEE} -6.0 & \ccol{FDF6F6} -3.1 & \ccol{F2F9F2} 2.0 & 0.9 & \ccol{ECF6EC} 3.1 & \ccol{F7FBF7} 1.2 & \ccol{F5FAF5} 1.5 & \ccol{E0F1E0} 4.9 & \ccol{85C885} 19.7 & \ccol{E2F2E2} 4.7 & \ccol{9CD29C} 16.1 \\*
 & XML & \ccol{FDF8F8} -2.4 & \ccol{FCF5F5} -3.6 & \ccol{FBEDED} -6.3 & \ccol{E3F2E3} 4.5 & \ccol{F4FAF4} 1.7 & \ccol{FBEEEE} -5.9 & \ccol{E5F3E5} 4.2 & \ccol{EFF8EF} 2.5 & \ccol{FCF5F5} -3.5 & \ccol{9BD29B} 16.2 & \ccol{E1F1E1} 4.9 & \ccol{B6DEB6} 11.9 \\*
 & Markdown & \ccol{E1F2E1} 4.7 & \ccol{FDF9F9} -2.0 & \ccol{F4FAF4} 1.7 & \ccol{E9F5E9} 3.5 & 0.8 & \ccol{FCF3F3} -4.3 & \ccol{B9DFB9} 11.3 & 0.8 & -0.7 & \ccol{8FCD8F} 18.2 & \ccol{E1F1E1} 4.9 & \ccol{CFE9CF} 7.7 \\*
 & LaTeX & \ccol{FCF1F1} -4.9 & \ccol{FDF6F6} -2.9 & \ccol{FCF3F3} -4.1 & \ccol{FDF8F8} -2.5 & \ccol{F8FCF8} 1.1 & \ccol{F8DFDF} -11.3 & \ccol{FBF0F0} -5.4 & 0.0 & \ccol{F0F8F0} 2.3 & \ccol{C7E6C7} 9.1 & \ccol{E1F1E1} 4.9 & \ccol{E1F1E1} 4.9 \\
\noalign{\vspace{0.7em}}\nopagebreak
\multirow{4}{*}{\texttt{·ET·}} & JSON & \ccol{FAE7E7} -8.4 & \ccol{FDF9F9} -2.0 & \ccol{E9F5E9} 3.5 & -0.5 & -0.3 & 0.3 & -0.3 & \ccol{EEF7EE} 2.6 & \ccol{EBF6EB} 3.1 & \ccol{B4DDB4} 12.1 & \ccol{E3F2E3} 4.5 & \ccol{98D198} 16.7 \\*
 & XML & 0.7 & \ccol{FDF6F6} -3.1 & \ccol{FAEBEB} -7.3 & \ccol{ECF6EC} 3.0 & \ccol{FEFBFB} -1.3 & 0.9 & \ccol{E8F4E8} 3.7 & \ccol{EEF7EE} 2.7 & 0.3 & \ccol{BDE1BD} 10.6 & \ccol{DCEFDC} 5.7 & \ccol{A8D8A8} 14.1 \\*
 & Markdown & \ccol{FDF9F9} -1.9 & -0.2 & \ccol{E4F3E4} 4.3 & \ccol{F8FCF8} 1.0 & 0.1 & \ccol{F8FCF8} 1.1 & -0.3 & \ccol{F7FBF7} 1.2 & -0.7 & \ccol{98D198} 16.7 & \ccol{DFF1DF} 5.1 & \ccol{9AD29A} 16.3 \\*
 & LaTeX & \ccol{FCF1F1} -4.9 & \ccol{FDF6F6} -3.1 & \ccol{E6F4E6} 3.9 & \ccol{F2F9F2} 2.0 & -0.3 & \ccol{FDF9F9} -2.1 & 0.2 & \ccol{F1F8F1} 2.3 & \ccol{DFF0DF} 5.1 & \ccol{9BD29B} 16.2 & \ccol{DDF0DD} 5.4 & \ccol{A2D5A2} 15.1 \\
\noalign{\vspace{0.7em}}\nopagebreak
\multirow{4}{*}{\texttt{GET·}} & JSON & \ccol{FAEAEA} -7.4 & \ccol{F8E0E0} -11.0 & \ccol{FCF3F3} -4.3 & \ccol{D6ECD6} 6.6 & \ccol{F8FCF8} 1.1 & \ccol{F1F8F1} 2.3 & \ccol{F7FBF7} 1.2 & \ccol{EEF7EE} 2.7 & \ccol{E0F1E0} 4.9 & \ccol{98D198} 16.7 & \ccol{E1F1E1} 4.9 & \ccol{A6D7A6} 14.5 \\*
 & XML & \ccol{F9E3E3} -9.9 & \ccol{F8E2E2} -10.4 & \ccol{FAE9E9} -7.9 & \ccol{E9F5E9} 3.5 & \ccol{F2F9F2} 2.1 & \ccol{F3F9F3} 1.9 & \ccol{EEF7EE} 2.7 & \ccol{EFF8EF} 2.5 & \ccol{F3F9F3} 1.9 & \ccol{95CF95} 17.2 & \ccol{E4F3E4} 4.3 & \ccol{99D199} 16.5 \\*
 & Markdown & \ccol{FCF4F4} -3.9 & \ccol{FCF4F4} -4.0 & -0.5 & \ccol{FEFCFC} -1.0 & \ccol{F4FAF4} 1.7 & 0.3 & \ccol{D5ECD5} 6.7 & 0.4 & \ccol{E0F1E0} 4.9 & \ccol{B1DCB1} 12.6 & \ccol{DFF1DF} 5.1 & \ccol{99D199} 16.5 \\*
 & LaTeX & \ccol{FBF0F0} -5.4 & \ccol{FBF0F0} -5.2 & -0.5 & \ccol{F5FAF5} 1.5 & -0.9 & \ccol{FDF9F9} -1.9 & \ccol{E1F2E1} 4.7 & 0.8 & \ccol{EFF8EF} 2.5 & \ccol{A8D8A8} 14.1 & \ccol{DCEFDC} 5.6 & \ccol{A2D5A2} 15.1 \\
\noalign{\vspace{0.7em}}\nopagebreak
\multirow{4}{*}{\texttt{G·TC}} & JSON & \ccol{F9E5E5} -9.4 & \ccol{F9E5E5} -9.4 & \ccol{FBEDED} -6.5 & \ccol{E6F3E6} 4.0 & \ccol{F7FBF7} 1.3 & \ccol{F5FAF5} 1.5 & \ccol{F4FAF4} 1.7 & \ccol{F2F9F2} 2.1 & \ccol{F0F8F0} 2.3 & \ccol{AEDAAE} 13.1 & \ccol{EFF8EF} 2.4 & \ccol{9ED39E} 15.7 \\*
 & XML & \ccol{FBEEEE} -5.9 & \ccol{F7DBDB} -13.0 & \ccol{FBEDED} -6.5 & \ccol{EFF8EF} 2.5 & 0.1 & \ccol{FDF8F8} -2.3 & \ccol{E8F4E8} 3.7 & -0.6 & 0.7 & \ccol{92CE92} 17.7 & \ccol{EAF6EA} 3.3 & \ccol{BDE1BD} 10.7 \\*
 & Markdown & \ccol{F1F8F1} 2.2 & \ccol{FCF5F5} -3.6 & \ccol{FDF8F8} -2.3 & \ccol{E3F2E3} 4.5 & \ccol{FEFBFB} -1.1 & \ccol{FDF6F6} -3.1 & \ccol{84C884} 22.4 & 0.6 & \ccol{FCF2F2} -4.5 & \ccol{9BD29B} 16.2 & \ccol{E8F4E8} 3.7 & \ccol{EAF6EA} 3.3 \\*
 & LaTeX & \ccol{FBF0F0} -5.4 & \ccol{FBEDED} -6.6 & \ccol{F3C7C7} -20.3 & \ccol{E9F5E9} 3.5 & -0.7 & \ccol{FDF7F7} -2.7 & \ccol{D5ECD5} 6.7 & \ccol{FEFBFB} -1.2 & \ccol{F2F9F2} 2.1 & \ccol{92CE92} 17.7 & \ccol{EEF7EE} 2.7 & \ccol{9ED39E} 15.7 \\
\noalign{\vspace{0.7em}}\nopagebreak
\multirow{4}{*}{\texttt{·ETC}} & JSON & \ccol{F8E1E1} -10.9 & \ccol{F9E3E3} -10.0 & \ccol{F9E5E5} -9.3 & \ccol{F2F9F2} 2.0 & \ccol{FEFAFA} -1.5 & 0.1 & \ccol{E1F2E1} 4.7 & 0.4 & \ccol{F0F8F0} 2.3 & \ccol{B1DCB1} 12.6 & \ccol{E3F2E3} 4.5 & \ccol{9FD49F} 15.5 \\*
 & XML & \ccol{FBEDED} -6.4 & \ccol{F9E2E2} -10.3 & \ccol{FCF2F2} -4.7 & \ccol{FDF9F9} -2.0 & -0.5 & 0.9 & -0.8 & 0.4 & 0.9 & \ccol{B1DCB1} 12.6 & \ccol{E4F3E4} 4.3 & \ccol{AFDBAF} 12.9 \\*
 & Markdown & \ccol{FCF4F4} -3.9 & \ccol{FCF4F4} -3.9 & \ccol{FCF4F4} -3.9 & 0.0 & -0.7 & \ccol{F1F8F1} 2.3 & \ccol{DBEFDB} 5.7 & 0.4 & \ccol{F8FBF8} 1.1 & \ccol{AEDAAE} 13.1 & \ccol{E4F3E4} 4.3 & \ccol{B2DCB2} 12.5 \\*
 & LaTeX & \ccol{FDF8F8} -2.4 & \ccol{FBF0F0} -5.4 & -0.5 & \ccol{E9F5E9} 3.5 & \ccol{FEFAFA} -1.7 & \ccol{FDF6F6} -2.9 & \ccol{E5F3E5} 4.2 & 0.8 & \ccol{EFF8EF} 2.5 & \ccol{B1DCB1} 12.6 & \ccol{F8FCF8} 1.1 & \ccol{9AD29A} 16.3 \\
\noalign{\vspace{0.7em}}\nopagebreak
\multirow{4}{*}{\texttt{GETC}} & JSON & \ccol{F8DEDE} -12.0 & \ccol{F9E3E3} -10.0 & \ccol{FCF4F4} -3.9 & \ccol{CAE7CA} 8.6 & 0.1 & \ccol{ECF6EC} 3.1 & \ccol{F7FBF7} 1.2 & 0.2 & \ccol{F2F9F2} 2.1 & \ccol{A8D8A8} 14.1 & \ccol{E8F4E8} 3.7 & \ccol{96D096} 17.1 \\*
 & XML & \ccol{F6D8D8} -14.0 & \ccol{F4CECE} -17.6 & \ccol{F9E2E2} -10.3 & \ccol{E6F3E6} 4.0 & -0.3 & 0.1 & \ccol{F4FAF4} 1.7 & -0.2 & \ccol{E0F1E0} 4.9 & \ccol{8CCB8C} 18.7 & \ccol{E9F5E9} 3.5 & \ccol{97D097} 16.9 \\*
 & Markdown & -0.8 & \ccol{FBEFEF} -5.8 & \ccol{F5FAF5} 1.5 & -0.5 & \ccol{FDF8F8} -2.3 & -0.3 & \ccol{B6DEB6} 11.8 & 0.4 & \ccol{E2F2E2} 4.7 & \ccol{A4D6A4} 14.6 & \ccol{DEF0DE} 5.3 & \ccol{ADDAAD} 13.3 \\*
 & LaTeX & \ccol{FDF5F5} -3.4 & \ccol{FAE8E8} -8.2 & -0.3 & \ccol{F8FCF8} 1.0 & 0.7 & \ccol{F4FAF4} 1.7 & \ccol{F1F8F1} 2.2 & 0.4 & \ccol{DFF0DF} 5.1 & \ccol{BAE0BA} 11.1 & \ccol{EAF6EA} 3.3 & \ccol{9FD49F} 15.5 \\
\noalign{\vspace{1em}}
\rowcolor{gray!25} \multicolumn{2}{l}{\hficon~\large\textbf{smollm3-3b}} & 22.9 & 67.4 & 38.4 & 29.6 & 89.0 & 38.3 & 22.9 & 67.4 & 38.4 & 29.6 & 89.0 & 38.3 \\
\multirow{4}{*}{\texttt{G·T·}} & JSON & \ccol{F3F9F3} 1.9 & \ccol{F6D5D5} -15.0 & \ccol{F8DFDF} -11.6 & \ccol{E5F3E5} 4.2 & \ccol{FEFBFB} -1.4 & \ccol{FDF6F6} -3.1 & \ccol{FBF0F0} -5.2 & \ccol{EEF7EE} 2.7 & -0.6 & \ccol{D5ECD5} 6.7 & 0.6 & \ccol{E1F1E1} 4.9 \\*
 & XML & 0.8 & \ccol{FCF4F4} -4.0 & \ccol{F7DBDB} -13.0 & \ccol{F4FAF4} 1.7 & \ccol{FEFAFA} -1.6 & \ccol{FCF2F2} -4.7 & \ccol{DAEEDA} 5.9 & -0.5 & -0.6 & \ccol{BCE1BC} 10.8 & 0.4 & \ccol{D7EDD7} 6.5 \\*
 & Markdown & \ccol{C8E6C8} 8.9 & -0.1 & \ccol{FAE9E9} -8.0 & \ccol{D2EBD2} 7.2 & \ccol{FEFAFA} -1.6 & \ccol{FCF3F3} -4.3 & \ccol{D4ECD4} 6.9 & \ccol{FEFCFC} -1.1 & \ccol{FDF8F8} -2.2 & \ccol{B9DFB9} 11.3 & \ccol{F6FBF6} 1.4 & \ccol{E4F3E4} 4.3 \\*
 & LaTeX & \ccol{BEE2BE} 10.4 & \ccol{FBEDED} -6.4 & \ccol{FAEBEB} -7.0 & \ccol{D5ECD5} 6.7 & \ccol{FDF7F7} -2.8 & \ccol{FDFAFA} -1.7 & \ccol{E7F4E7} 3.9 & 0.2 & \ccol{FEFBFB} -1.4 & \ccol{E8F4E8} 3.7 & 0.4 & \ccol{E4F3E4} 4.3 \\
\noalign{\vspace{0.7em}}\nopagebreak
\multirow{4}{*}{\texttt{·ET·}} & JSON & \ccol{FBEFEF} -5.7 & \ccol{FCF3F3} -4.0 & \ccol{FDF6F6} -3.2 & \ccol{F4FAF4} 1.7 & \ccol{FEFAFA} -1.6 & \ccol{FDF9F9} -1.9 & \ccol{EAF5EA} 3.4 & 0.3 & \ccol{EFF7EF} 2.6 & \ccol{E5F3E5} 4.2 & -0.2 & \ccol{A2D5A2} 15.1 \\*
 & XML & \ccol{FEFBFB} -1.2 & \ccol{FCF1F1} -4.8 & \ccol{FBEDED} -6.4 & \ccol{E2F2E2} 4.7 & -0.8 & \ccol{FEFBFB} -1.3 & 0.3 & \ccol{F0F8F0} 2.3 & \ccol{F1F8F1} 2.2 & \ccol{E2F2E2} 4.7 & -0.0 & \ccol{B8DFB8} 11.5 \\*
 & Markdown & \ccol{FDF9F9} -2.2 & -0.8 & \ccol{FCF2F2} -4.6 & \ccol{EEF7EE} 2.7 & \ccol{FEFBFB} -1.4 & \ccol{E8F4E8} 3.7 & \ccol{E4F2E4} 4.4 & 0.1 & \ccol{E8F5E8} 3.6 & \ccol{BCE1BC} 10.8 & 0.2 & \ccol{B9E0B9} 11.3 \\*
 & LaTeX & \ccol{F0F8F0} 2.4 & \ccol{FCF4F4} -3.7 & -0.8 & \ccol{E5F3E5} 4.2 & -0.8 & -0.9 & \ccol{E0F1E0} 4.9 & 0.1 & 0.6 & \ccol{DBEFDB} 5.7 & 0.2 & \ccol{B3DDB3} 12.3 \\
\noalign{\vspace{0.7em}}\nopagebreak
\multirow{4}{*}{\texttt{GET·}} & JSON & \ccol{E0F1E0} 4.9 & \ccol{F5D1D1} -16.6 & \ccol{FDF5F5} -3.4 & \ccol{F1F8F1} 2.2 & \ccol{FDF7F7} -2.6 & -0.1 & \ccol{EDF7ED} 2.9 & 0.2 & 0.8 & \ccol{D5ECD5} 6.7 & \ccol{FDF8F8} -2.4 & \ccol{B8DFB8} 11.5 \\*
 & XML & \ccol{DDF0DD} 5.4 & \ccol{F9E6E6} -9.1 & \ccol{F9E4E4} -9.6 & \ccol{FDF5F5} -3.4 & \ccol{FDF6F6} -3.3 & \ccol{FEFBFB} -1.3 & \ccol{F0F8F0} 2.4 & 0.3 & \ccol{F1F8F1} 2.2 & \ccol{D5ECD5} 6.7 & \ccol{FBEEEE} -6.2 & \ccol{B6DEB6} 11.9 \\*
 & Markdown & \ccol{F3F9F3} 1.9 & \ccol{FDF6F6} -3.2 & \ccol{FBF0F0} -5.4 & \ccol{EBF6EB} 3.2 & \ccol{FEFBFB} -1.4 & \ccol{FCF0F0} -5.1 & \ccol{D4ECD4} 6.9 & 0.3 & \ccol{FEFCFC} -1.0 & \ccol{D8EDD8} 6.2 & 1.0 & \ccol{D1EAD1} 7.5 \\*
 & LaTeX & \ccol{EAF5EA} 3.4 & \ccol{F9E3E3} -10.2 & \ccol{FCF5F5} -3.6 & -0.8 & \ccol{FEFBFB} -1.2 & \ccol{FDF9F9} -2.1 & \ccol{EAF5EA} 3.4 & 0.5 & \ccol{F0F8F0} 2.4 & \ccol{D2EBD2} 7.2 & -0.6 & \ccol{A7D7A7} 14.3 \\
\noalign{\vspace{0.7em}}\nopagebreak
\multirow{4}{*}{\texttt{G·TC}} & JSON & \ccol{CEE9CE} 7.9 & \ccol{F2C2C2} -22.2 & \ccol{FDF9F9} -2.0 & \ccol{EBF6EB} 3.2 & \ccol{FDF7F7} -2.8 & \ccol{E9F5E9} 3.5 & \ccol{DAEEDA} 5.9 & \ccol{F5FAF5} 1.5 & 1.0 & \ccol{D2EBD2} 7.2 & -0.6 & \ccol{C7E6C7} 9.1 \\*
 & XML & \ccol{FDF6F6} -3.2 & \ccol{F7DBDB} -13.1 & \ccol{FAE9E9} -7.8 & \ccol{D2EBD2} 7.2 & \ccol{FDF9F9} -2.0 & \ccol{FDF8F8} -2.5 & \ccol{E0F1E0} 4.9 & \ccol{FDF9F9} -2.0 & \ccol{F5FAF5} 1.6 & \ccol{BFE2BF} 10.3 & \ccol{FCF3F3} -4.0 & \ccol{CEE9CE} 7.9 \\*
 & Markdown & \ccol{D1EAD1} 7.4 & \ccol{FDF6F6} -2.9 & \ccol{F9E5E5} -9.2 & 0.7 & \ccol{FDF8F8} -2.4 & \ccol{FDF6F6} -2.9 & \ccol{C1E3C1} 9.9 & -0.7 & -0.0 & \ccol{91CD91} 17.8 & 0.6 & \ccol{EEF7EE} 2.7 \\*
 & LaTeX & \ccol{99D199} 16.5 & \ccol{F8DEDE} -11.9 & \ccol{FAEAEA} -7.6 & \ccol{D5ECD5} 6.7 & \ccol{F8E2E2} -10.6 & -0.9 & \ccol{C8E6C8} 8.9 & \ccol{FBF0F0} -5.4 & \ccol{F2F9F2} 2.0 & \ccol{B6DEB6} 11.8 & \ccol{FBEEEE} -6.0 & \ccol{C7E6C7} 9.1 \\
\noalign{\vspace{0.7em}}\nopagebreak
\multirow{4}{*}{\texttt{·ETC}} & JSON & \ccol{FDF7F7} -2.7 & \ccol{F0BCBC} -24.2 & \ccol{FBF0F0} -5.2 & \ccol{F1F8F1} 2.2 & \ccol{FCF1F1} -5.0 & \ccol{F8FCF8} 1.1 & \ccol{E7F4E7} 3.9 & 0.1 & 0.4 & \ccol{E5F3E5} 4.2 & \ccol{FCF3F3} -4.2 & \ccol{A7D7A7} 14.3 \\*
 & XML & \ccol{FDF6F6} -3.2 & \ccol{F7DADA} -13.5 & \ccol{FCF3F3} -4.2 & \ccol{FEFBFB} -1.4 & \ccol{FCF3F3} -4.4 & \ccol{FEFAFA} -1.5 & -0.7 & -0.5 & \ccol{F0F8F0} 2.4 & 0.2 & \ccol{FAEAEA} -7.4 & \ccol{CFE9CF} 7.7 \\*
 & Markdown & \ccol{FEFBFB} -1.2 & \ccol{FDF8F8} -2.2 & \ccol{FCF3F3} -4.2 & \ccol{F7FBF7} 1.2 & \ccol{FCF4F4} -3.8 & -0.3 & \ccol{EAF5EA} 3.4 & -0.9 & \ccol{F0F8F0} 2.4 & \ccol{E2F2E2} 4.7 & -0.2 & \ccol{ADDAAD} 13.3 \\*
 & LaTeX & \ccol{CBE7CB} 8.4 & \ccol{FAEAEA} -7.4 & -0.0 & \ccol{F1F8F1} 2.2 & \ccol{FBEDED} -6.4 & \ccol{FDF9F9} -2.1 & \ccol{D1EAD1} 7.4 & \ccol{FEFCFC} -1.1 & \ccol{EDF7ED} 2.8 & \ccol{C9E6C9} 8.7 & \ccol{FEFBFB} -1.4 & \ccol{8DCC8D} 18.5 \\
\noalign{\vspace{0.7em}}\nopagebreak
\multirow{4}{*}{\texttt{GETC}} & JSON & \ccol{C1E3C1} 9.9 & \ccol{F2C2C2} -21.9 & \ccol{FBECEC} -6.6 & \ccol{E5F3E5} 4.2 & \ccol{FCF1F1} -4.8 & \ccol{F5FAF5} 1.5 & \ccol{E7F4E7} 3.9 & 0.2 & \ccol{F1F8F1} 2.2 & \ccol{E5F3E5} 4.2 & \ccol{F9E6E6} -8.8 & \ccol{C2E3C2} 9.9 \\*
 & XML & \ccol{FEFBFB} -1.2 & \ccol{F7DADA} -13.5 & \ccol{FAE7E7} -8.6 & \ccol{F7FBF7} 1.2 & \ccol{FBF0F0} -5.4 & \ccol{F4FAF4} 1.7 & \ccol{FEFAFA} -1.7 & \ccol{FCF3F3} -4.2 & 0.2 & \ccol{EBF6EB} 3.2 & \ccol{F7D9D9} -13.6 & \ccol{BCE1BC} 10.9 \\*
 & Markdown & \ccol{EDF7ED} 2.9 & \ccol{FBEEEE} -6.1 & \ccol{FCF4F4} -3.8 & \ccol{F7FBF7} 1.2 & \ccol{FBF0F0} -5.5 & 0.3 & 0.3 & -0.1 & 0.2 & \ccol{D5ECD5} 6.7 & -0.2 & \ccol{C4E4C4} 9.5 \\*
 & LaTeX & \ccol{F6FBF6} 1.3 & \ccol{F6D7D7} -14.3 & \ccol{FDFAFA} -1.8 & \ccol{F4FAF4} 1.7 & \ccol{FBEEEE} -6.2 & 0.5 & \ccol{DAEEDA} 5.9 & \ccol{FDF9F9} -2.0 & \ccol{E8F5E8} 3.6 & \ccol{B9DFB9} 11.3 & \ccol{FDF8F8} -2.4 & \ccol{C4E4C4} 9.5 \\
\noalign{\vspace{1em}}
\rowcolor{gray!25} \multicolumn{2}{l}{\nvidiaicon~\large\textbf{nemotron3-nano}} & 53.9 & 71.7 & 43.1 & 54.0 & 91.6 & 60.3 & 53.9 & 71.7 & 43.1 & 54.0 & 91.6 & 60.3 \\
\multirow{4}{*}{\texttt{G·T·}} & JSON & \ccol{F7DBDB} -13.0 & 0.7 & \ccol{C6E5C6} 9.3 & \ccol{FEFCFC} -1.0 & 0.0 & \ccol{C1E3C1} 10.1 & \ccol{FAE9E9} -7.9 & \ccol{E9F5E9} 3.5 & \ccol{CFE9CF} 7.7 & \ccol{ECF6EC} 3.0 & \ccol{EDF7ED} 2.8 & \ccol{A6D7A6} 14.5 \\*
 & XML & \ccol{DBEFDB} 5.7 & \ccol{FEFBFB} -1.4 & 0.9 & \ccol{DCEFDC} 5.6 & 0.2 & \ccol{AAD9AA} 13.7 & \ccol{FCF4F4} -3.9 & \ccol{F1F9F1} 2.1 & \ccol{CAE7CA} 8.5 & \ccol{C3E4C3} 9.6 & \ccol{EFF7EF} 2.6 & \ccol{B4DDB4} 12.1 \\*
 & Markdown & \ccol{DEF0DE} 5.2 & \ccol{F7DDDD} -12.2 & \ccol{F8E1E1} -10.9 & \ccol{C3E4C3} 9.6 & -0.4 & \ccol{BFE2BF} 10.3 & \ccol{EBF6EB} 3.2 & \ccol{F7FBF7} 1.3 & \ccol{FEFAFA} -1.5 & \ccol{A4D6A4} 14.6 & -0.8 & \ccol{E8F4E8} 3.7 \\*
 & LaTeX & \ccol{F6D8D8} -14.0 & \ccol{EBA4A4} -33.0 & \ccol{FAECEC} -6.9 & \ccol{B4DDB4} 12.1 & \ccol{FEFAFA} -1.6 & \ccol{CDE8CD} 8.1 & \ccol{F9E5E5} -9.4 & 0.4 & \ccol{FDF6F6} -2.9 & \ccol{AEDAAE} 13.1 & -0.2 & \ccol{D6ECD6} 6.7 \\
\noalign{\vspace{0.7em}}\nopagebreak
\multirow{4}{*}{\texttt{·ET·}} & JSON & \ccol{F7DCDC} -12.5 & \ccol{FCF1F1} -5.0 & \ccol{D6ECD6} 6.7 & \ccol{E9F5E9} 3.5 & \ccol{FDFAFA} -1.8 & -0.1 & \ccol{F7DADA} -13.5 & \ccol{F3F9F3} 1.9 & \ccol{CEE9CE} 7.9 & \ccol{84C884} 24.7 & \ccol{FBEFEF} -5.7 & \ccol{A3D6A3} 14.9 \\*
 & XML & \ccol{F1F8F1} 2.2 & \ccol{E58484} -44.9 & \ccol{84C884} 20.5 & \ccol{FDF8F8} -2.5 & 0.0 & \ccol{F4FAF4} 1.7 & \ccol{F8DEDE} -12.0 & \ccol{F1F9F1} 2.1 & \ccol{D1EAD1} 7.5 & \ccol{A8D8A8} 14.1 & \ccol{FCF1F1} -4.8 & \ccol{A6D7A6} 14.5 \\*
 & Markdown & \ccol{F7DBDB} -13.0 & \ccol{E4F2E4} 4.4 & \ccol{FCF3F3} -4.1 & \ccol{FDF6F6} -3.0 & 0.4 & \ccol{E8F4E8} 3.7 & \ccol{F8E1E1} -10.9 & \ccol{EDF7ED} 2.8 & \ccol{C8E6C8} 8.9 & \ccol{C7E6C7} 9.1 & \ccol{FAEAEA} -7.6 & \ccol{B1DCB1} 12.7 \\*
 & LaTeX & \ccol{84C884} 21.4 & \ccol{EA9D9D} -35.7 & \ccol{FBF0F0} -5.3 & \ccol{F5FAF5} 1.5 & -0.2 & \ccol{F8FCF8} 1.1 & \ccol{F8E2E2} -10.4 & \ccol{F6FBF6} 1.3 & \ccol{E7F4E7} 3.9 & \ccol{84C884} 21.7 & \ccol{FCF3F3} -4.2 & \ccol{D2EBD2} 7.3 \\
\noalign{\vspace{0.7em}}\nopagebreak
\multirow{4}{*}{\texttt{GET·}} & JSON & \ccol{F5D0D0} -17.0 & \ccol{F5FAF5} 1.5 & \ccol{C6E5C6} 9.3 & \ccol{D6ECD6} 6.6 & \ccol{F7FBF7} 1.2 & \ccol{C1E3C1} 10.1 & \ccol{F9E6E6} -8.9 & \ccol{EDF7ED} 2.8 & \ccol{CDE8CD} 8.1 & \ccol{D3EBD3} 7.1 & \ccol{EDF7ED} 2.8 & \ccol{B1DCB1} 12.7 \\*
 & XML & \ccol{FDF5F5} -3.4 & \ccol{E8F4E8} 3.7 & \ccol{FCF3F3} -4.1 & \ccol{DCEFDC} 5.6 & -0.0 & \ccol{ADDAAD} 13.3 & \ccol{FBEEEE} -5.9 & \ccol{EBF6EB} 3.2 & \ccol{E2F2E2} 4.7 & \ccol{BAE0BA} 11.1 & \ccol{ECF6EC} 3.0 & \ccol{BDE1BD} 10.7 \\*
 & Markdown & \ccol{F9E3E3} -9.9 & \ccol{EEF7EE} 2.7 & \ccol{FCF1F1} -4.9 & \ccol{ECF6EC} 3.0 & 0.4 & \ccol{AFDBAF} 12.9 & \ccol{FAE7E7} -8.4 & \ccol{EEF7EE} 2.6 & \ccol{D1EAD1} 7.5 & \ccol{DFF1DF} 5.0 & \ccol{FEFBFB} -1.2 & \ccol{D3EBD3} 7.1 \\*
 & LaTeX & \ccol{F7DBDB} -13.0 & \ccol{FBF0F0} -5.3 & \ccol{FDFAFA} -1.7 & \ccol{DCEFDC} 5.6 & -0.6 & \ccol{B7DEB7} 11.7 & \ccol{F9E3E3} -9.9 & \ccol{FEFAFA} -1.6 & 0.7 & \ccol{D0EAD0} 7.6 & \ccol{ECF6EC} 3.0 & \ccol{D4ECD4} 6.9 \\
\noalign{\vspace{0.7em}}\nopagebreak
\multirow{4}{*}{\texttt{G·TC}} & JSON & \ccol{F8DEDE} -12.0 & 0.5 & \ccol{C4E4C4} 9.5 & \ccol{E6F3E6} 4.0 & \ccol{F5FAF5} 1.6 & \ccol{C2E3C2} 9.9 & \ccol{FAEAEA} -7.4 & \ccol{F1F8F1} 2.2 & \ccol{CFE9CF} 7.7 & \ccol{CAE7CA} 8.6 & \ccol{E8F5E8} 3.6 & \ccol{A3D6A3} 14.9 \\*
 & XML & \ccol{DBEFDB} 5.7 & \ccol{FCF2F2} -4.5 & \ccol{EFF8EF} 2.5 & \ccol{C3E4C3} 9.6 & 1.0 & \ccol{BCE1BC} 10.9 & \ccol{D8EDD8} 6.2 & 0.3 & \ccol{CAE7CA} 8.5 & \ccol{ABD9AB} 13.6 & 0.8 & \ccol{B7DEB7} 11.7 \\*
 & Markdown & \ccol{CCE8CC} 8.2 & \ccol{FEFAFA} -1.5 & \ccol{F9E3E3} -10.1 & \ccol{DFF1DF} 5.0 & \ccol{FEFCFC} -1.0 & \ccol{CDE8CD} 8.1 & \ccol{C6E5C6} 9.3 & 0.1 & \ccol{FDFAFA} -1.7 & \ccol{84C884} 25.2 & 0.9 & \ccol{F7DBDB} -12.9 \\*
 & LaTeX & \ccol{FAE9E9} -7.9 & \ccol{F9E5E5} -9.4 & \ccol{FCF4F4} -3.9 & \ccol{B4DDB4} 12.1 & \ccol{FDF7F7} -2.6 & \ccol{B8DFB8} 11.5 & \ccol{FBEDED} -6.4 & \ccol{FCF5F5} -3.5 & \ccol{D3EBD3} 7.1 & \ccol{C7E6C7} 9.1 & \ccol{FDF7F7} -2.8 & \ccol{D7EDD7} 6.5 \\
\noalign{\vspace{0.7em}}\nopagebreak
\multirow{4}{*}{\texttt{·ETC}} & JSON & \ccol{F8DFDF} -11.5 & \ccol{FDF5F5} -3.3 & \ccol{E8F4E8} 3.7 & \ccol{E9F5E9} 3.5 & -0.2 & \ccol{EAF6EA} 3.3 & \ccol{F9E3E3} -9.9 & \ccol{F3F9F3} 1.9 & \ccol{D9EED9} 6.1 & \ccol{D9EED9} 6.1 & \ccol{F0F8F0} 2.4 & \ccol{A6D7A6} 14.5 \\*
 & XML & \ccol{F0B8B8} -25.6 & \ccol{FDF9F9} -1.9 & \ccol{C9E7C9} 8.7 & \ccol{F5FAF5} 1.5 & 0.2 & \ccol{E3F2E3} 4.5 & \ccol{F4CFCF} -17.5 & \ccol{E3F2E3} 4.4 & \ccol{D4ECD4} 6.9 & \ccol{F5FAF5} 1.5 & \ccol{EBF6EB} 3.2 & \ccol{A8D8A8} 14.1 \\*
 & Markdown & \ccol{F2C3C3} -21.6 & \ccol{FDF6F6} -3.2 & \ccol{FCF2F2} -4.5 & -0.0 & \ccol{FDF7F7} -2.6 & \ccol{EAF6EA} 3.3 & \ccol{F7DADA} -13.5 & \ccol{FDFAFA} -1.7 & \ccol{F5FAF5} 1.5 & -0.0 & \ccol{FCF3F3} -4.2 & \ccol{E5F3E5} 4.1 \\*
 & LaTeX & \ccol{F3C8C8} -20.0 & \ccol{E3F2E3} 4.5 & \ccol{F5FAF5} 1.5 & \ccol{FDF9F9} -2.0 & \ccol{FEFAFA} -1.6 & \ccol{F8FCF8} 1.1 & \ccol{FAE7E7} -8.4 & \ccol{FDF5F5} -3.3 & \ccol{D9EED9} 6.1 & \ccol{EFF8EF} 2.5 & -0.8 & \ccol{B4DDB4} 12.1 \\
\noalign{\vspace{0.7em}}\nopagebreak
\multirow{4}{*}{\texttt{GETC}} & JSON & \ccol{F2C5C5} -21.0 & 0.0 & \ccol{D7EDD7} 6.5 & \ccol{E3F2E3} 4.5 & \ccol{FEFBFB} -1.4 & \ccol{B7DEB7} 11.7 & \ccol{F9E5E5} -9.4 & \ccol{EDF7ED} 2.8 & \ccol{CDE8CD} 8.1 & \ccol{ECF6EC} 3.0 & \ccol{F6FBF6} 1.4 & \ccol{AFDBAF} 12.9 \\*
 & XML & \ccol{F9E6E6} -8.9 & \ccol{FEFAFA} -1.5 & \ccol{F3F9F3} 1.9 & \ccol{DFF1DF} 5.0 & \ccol{F6FBF6} 1.4 & \ccol{AFDBAF} 12.9 & \ccol{FCF2F2} -4.4 & 0.7 & \ccol{C9E7C9} 8.7 & \ccol{CDE8CD} 8.1 & \ccol{F1F8F1} 2.2 & \ccol{B2DCB2} 12.5 \\*
 & Markdown & \ccol{F9E6E6} -8.9 & 0.3 & \ccol{FCF3F3} -4.3 & \ccol{DCEFDC} 5.6 & 0.2 & \ccol{AEDBAE} 13.1 & \ccol{F9E5E5} -9.4 & \ccol{FDF8F8} -2.5 & \ccol{EEF7EE} 2.7 & \ccol{D9EED9} 6.1 & \ccol{FDF9F9} -2.0 & \ccol{FDFAFA} -1.7 \\*
 & LaTeX & \ccol{F8DFDF} -11.5 & \ccol{FAEAEA} -7.5 & \ccol{FDF6F6} -3.1 & \ccol{E9F5E9} 3.5 & -0.0 & \ccol{BCE1BC} 10.9 & \ccol{FAEAEA} -7.4 & \ccol{FDF5F5} -3.4 & \ccol{D1EAD1} 7.5 & \ccol{D9EED9} 6.1 & \ccol{F0F8F0} 2.4 & \ccol{CCE8CC} 8.3 \\
\end{longtable}
\setlength{\tabcolsep}{6pt}
\normalsize

\FloatBarrier

\section{Experimental Details}
\label{sec:appendix_details}

\begin{table*}[t]
\centering
\small
\resizebox{\textwidth}{!}{%
\begin{tabular}{@{}llp{7.2cm}@{}}
\toprule
\textbf{Parameter} & \textbf{Value} & \textbf{Notes} \\
\midrule
\multicolumn{3}{@{}l}{\textit{Generation Parameters}} \\[2pt]
Temperature          & 1.0   & \\
Top-$p$              & 1.0   & No nucleus filtering; applied to vLLM only (not supported by OpenAI/Anthropic APIs) \\
Max output tokens    & 8{,}192 & Per generation turn \\
Samples per question ($n$) & 1 & Single sample per question \\
Random seed          & 42    & Fixed for vLLM experiments; API providers use server-side randomness \\
Few-shot count       & 2     & In-context examples (when few-shot is enabled) \\
\midrule
\multicolumn{3}{@{}l}{\textit{Thinking Mode Mapping by Provider}} \\[2pt]
\multicolumn{3}{@{}l}{\quad\textbf{vLLM} (local models): binary only---OFF or HIGH via \texttt{enable\_thinking} template parameter} \\
\multicolumn{3}{@{}l}{\quad\textbf{OpenAI}: \texttt{reasoning\_effort} $\in$ \{none/minimal, low, medium, high\}; no temperature or top-$p$ support} \\[2pt]
\midrule
\multicolumn{3}{@{}l}{\textit{Decoupled (Two-Turn) Strategy}} \\[2pt]
Turn 1 temperature   & 1.0   & Same as generation default; inherited from config \\
Turn 2 temperature   & 1.0   & Reformatting turn; set per-turn in manifest \\
Turn 2 thinking      & OFF   & No reasoning during reformatting \\
Turn 2 samples       & 1     & Single reformat per reasoning output \\
\midrule
\multicolumn{3}{@{}l}{\textit{Infrastructure}} \\[2pt]
vLLM structured output backend & \texttt{llguidance} & Grammar-constrained decoding engine; \texttt{disable\_fallback=True} \\
Max examples per task & 500  & Default cap for tractable evaluation \\
\bottomrule
\end{tabular}}%
\caption{Default hyperparameters used across all experiments unless otherwise noted.}
\label{tab:hyperparameters}
\end{table*}

\FloatBarrier

\end{document}